\documentclass[sigconf, arxiv]{acmart}
\usepackage{booktabs}
\usepackage{multirow}
\usepackage{siunitx}
\AtBeginDocument{%
  \providecommand\BibTeX{{%
    \normalfont B\kern-0.5em{\scshape i\kern-0.25em b}\kern-0.8em\TeX}}}

\newif\iffinal
\finaltrue

\iffinal
  \newcommand\sutanay[1]{}
  \newcommand\logan[1]{}
  \newcommand\jenna[1]{}
  \newcommand\ganesh[1]{}
  \newcommand\neeraj[1]{}
  \newcommand\ian[1]{}
\else
  \newcommand\sutanay[1]{{\color{blue}[Sutanay: #1]}}
  \newcommand\logan[1]{{\color{violet}[Logan: #1]}}
  \newcommand\jenna[1]{{\color{green}[Jenna: #1]}}
  \newcommand\ganesh[1]{{\color{red}[Ganesh: #1]}}
  \newcommand\neeraj[1]{{\color{cyan}[Neeraj: #1]}}
  \newcommand\ian[1]{{\color{orange}[Ian: #1]}}
\fi

\newcommand\icfifty{IC\textsubscript{50}}

\begin{document}

\author{Logan Ward}
\affiliation{%
  \institution{Argonne National Laboratory}
}
\email{lward@anl.gov}

\author{Jenna A. Bilbrey}
\affiliation{%
  \institution{Pacific Northwest National Laboratory}
}
\email{jenna.pope@pnnl.gov}

\author{Sutanay Choudhury}
\affiliation{%
  \institution{Pacific Northwest National Laboratory}
}
\email{sutanay.choudhury@pnnl.gov}

\author{Neeraj Kumar}
\affiliation{%
  \institution{Pacific Northwest National Laboratory}
}
\email{neeraj.kumar@pnnl.gov}

\author{Ganesh Sivaraman}
\affiliation{%
  \institution{Argonne National Laboratory}
}
\email{gsivaraman@anl.gov}

%\title{Benchmarking of Graph Generative Models for Generation of Small Drug Molecules Optimized for Inhibitory Concentration}
% \title{Evening the Score: Tailoring the Scoring Function in Graph Generative Models of Small-Molecule Anti-SARS Drugs}
% \title{Automated Generation of Small Drug Molecules for SARS-CoV-2: A Graph Generative Model Perspective}
\title{Benchmarking Deep Graph Generative Models for Optimizing New Drug Molecules for COVID-19}
\renewcommand{\shortauthors}{Ward, et al.}
\settopmatter{printacmref=false}
\setcopyright{none}
\renewcommand\footnotetextcopyrightpermission[1]{}
\pagestyle{plain}
\begin{abstract}
 Design of new drug compounds with target properties is a key area of research in generative modeling. We present a small drug molecule design pipeline based on graph-generative models and a comparison study of two state-of-the-art graph generative models for designing COVID-19 targeted drug candidates: 1) a variational autoencoder-based approach (VAE) that uses prior knowledge of molecules that have been shown to be effective for earlier coronavirus treatments and 2) a deep Q-learning method (DQN) that generates optimized molecules without any proximity constraints. We evaluate the novelty of the automated molecule generation approaches by validating the candidate molecules with drug-protein binding affinity models. The VAE method produced two novel molecules with similar structures to the antiretroviral protease inhibitor Indinavir that show potential binding affinity for the SARS-CoV-2 protein target 3-chymotrypsin-like protease (3CL-protease).
%Our study also underscores the critical role attention-based graph neural networks play as a common component for both models. 
  
%   General story: VAE vs RL for drug design: benefits and tradeoffs offor drug design
%     - VAE: Using both the training data distribution and surrogate model
%     - RL; Using only the surrogate model, but free from other constraints
%     - Validation by scoring functions optimizing for IC50, QED etc.
%  - Difficulties in RL, VAE and MPNN for antiviral drug design?
%     - What did we have to do to make them work for COVID
%         - RL + VAE: Worked out of the box. 
%         - MPNN: Careful selection of read-out functions (do an experiment to show the benefits of attentions)
%  - Describing our validation methods (PDBA?)
\end{abstract}

\maketitle

\section{Introduction}

Deep learning has demonstrated the potential to revolutionize drug design by reducing the initial search space in the early stages of discovery \cite{GCPN, Li2018GraphGen, huang2020deeppurpose, COVID-repurpose-nature}.\ian{This could be clearer if you start by defining the drug design problem (finding molecules with specific properties) and the large size of the space (currently only in second paragraph). Also, it seems confusing to me to talk about a search space (first sentence) and molecule generation (second sentence). Is the problem one of search or generation?} By applying the appropriate algorithm trained on the appropriate data, novel molecules can be generated with target properties \cite{horwood2020molecular, khemchandani2020deepgraphmol, IBM,selfiegeneration2020}. Here, we evaluate methods for the high-performance intelligent design of small-molecule drug candidates with anti-SARS activity, with a specific focus on SARS-CoV-2, otherwise known as COVID-19. However, discovering potential lead candidates for COVID-19 presents a challenge to the scientific community due to the long timescale of the drug development process. There is a need to accelerate the design process through AI-driven workflows for effective drug compound development. \ian{Or: ``AI-driven workflows are a promising approach to accelerating the design of effective drug compounds.''}

The potential drug space is composed of over $10^{20}$ compounds, and candidates with suitable activity against specific proteins only narrows the search space to $10^4-10^5$ structures\ian{Meaning unclear: do you mean, ``even if we restrict ourselves to candidates that have been shown experimentally to demonstrate suitable activity against specific compounds, ...''?}.  Machine learning (ML) methods are actively used in this search-and-screen process. Candidate molecules generated by ML methods are passed to downstream verification via virtual high-throughput drug-protein binding techniques, drug synthesis, biological assay, and finally clinical trials \cite{dror2012biomolecular, Batra2020Covid, SummitDocking}.

Heterogeneous graphs provide a natural representation for small-molecule organic compounds, with nodes representing atoms in the molecular structure and edges representing bonds between the atoms \cite{weininger1990smiles}. This approach motivated the exploration of graph-generative models such as graph convolutional policy networks \cite{GCPN}, variational autoencoders \cite{JTVAE, samanta2019nevae, JTVAE-multi}, and variants of deep reinforcement learning \cite{MolDQN, staahl2019deep} for the target-driven optimization of drug molecules. The drug-molecule design task is defined as generating a set of graphs $G_{opt}$ such that for each graph $g \in G_{opt}$, $f_{opt}(g) \geq \delta$, where $f_{opt}$ is a property optimization function and $\delta$ is a user-specified threshold. Most methods optimize target molecules for properties that can be derived from the molecular structure, such as the octanol-water partition coefficient (logP) \cite{avdeef1998ph} and quantitative estimate of druglikeness (QED) \cite{bickerton2012quantifying}. 
% Such properties are shown to be powerful predictors of important factors such as oral bioavailability \cite{yoshida2000qsar}, and the selection of molecules with high druglikeness helps to reduce attrition during drug development \cite{DruglikeInfluence, QED}.  Given a graph $G = (V, E)$ where $V$ represents the set of nodes and $E$ is the set of edges, 

The focus of our work is two-fold to generate compounds for drug discovery, specifically for SARS-CoV-2. \ian{Reword? ``In the work reported here, we develop, evaluate, and compare two approaches to generating candidate drug compounds, specifically for SARS-CoV-2.''} In the first \textit{similarity approach}, a junction-tree based variational autoencoder (JT-VAE) \cite{JTVAE} is trained on a database of known drug molecules that has pIC\textsubscript{50} activity \cite{de2016sars}. The trained VAE is used to generate novel molecules optimized towards specific properties using Bayesian optimization (BO).  The second approach extends a graph-based deep reinforcement learning (DQN) method \cite{MolDQN} to generate molecules that are not constrained by their proximity to past anti-SARS compounds. For comparison, we use the same set of property scoring functions ($f_{opt}$) as the respective optimization and reward functions for benchmarking the JT-VAE and DQN approaches. 

\textbf{Contributions.} 1) The goal of our study is to perform multi-objective optimization for generating molecular structures by considering critical bioactivity properties that are typically not considered in structure-activity relationship approaches. Specifically, we focus on pIC\textsubscript{50} \cite{sebaugh2011guidelines}\ian{Confusing that you say multi-objective optimization, but then indicate just one property.}, which captures the potency of the drug towards a protease target and cannot be accurately estimated from the chemical structure \cite{IC50-QM}. 2) In this context, we assembled a new protease dataset with molecules that are active against various enzymatic assays. This is considered to be one\ian{what is ``one'' referring to here?} of the key properties while generating new drug molecules.  These molecules are filtered from experimental pharmacology data in CheMBL, BindingDB, and ToxCat \cite{bento2014chembl}. 3) Finally, we validate all top-ranking molecules against a Drug Target Binding Affinity (DBTA) classifier \cite{huang2020deeppurpose} to asses potential anti-SARS-CoV-2 activity. Full details of our implementation and source code are available at \textbf{https://github.com/exalearn/covid-drug-design}.
\ian{The ``contributions'' summarizes what you did, but not why someone should care.}
% included in our GitHub page (URL excluded to respect double-blind reviewing).

\begin{figure*}[!h] \centering
  \includegraphics[width=0.8\textwidth]{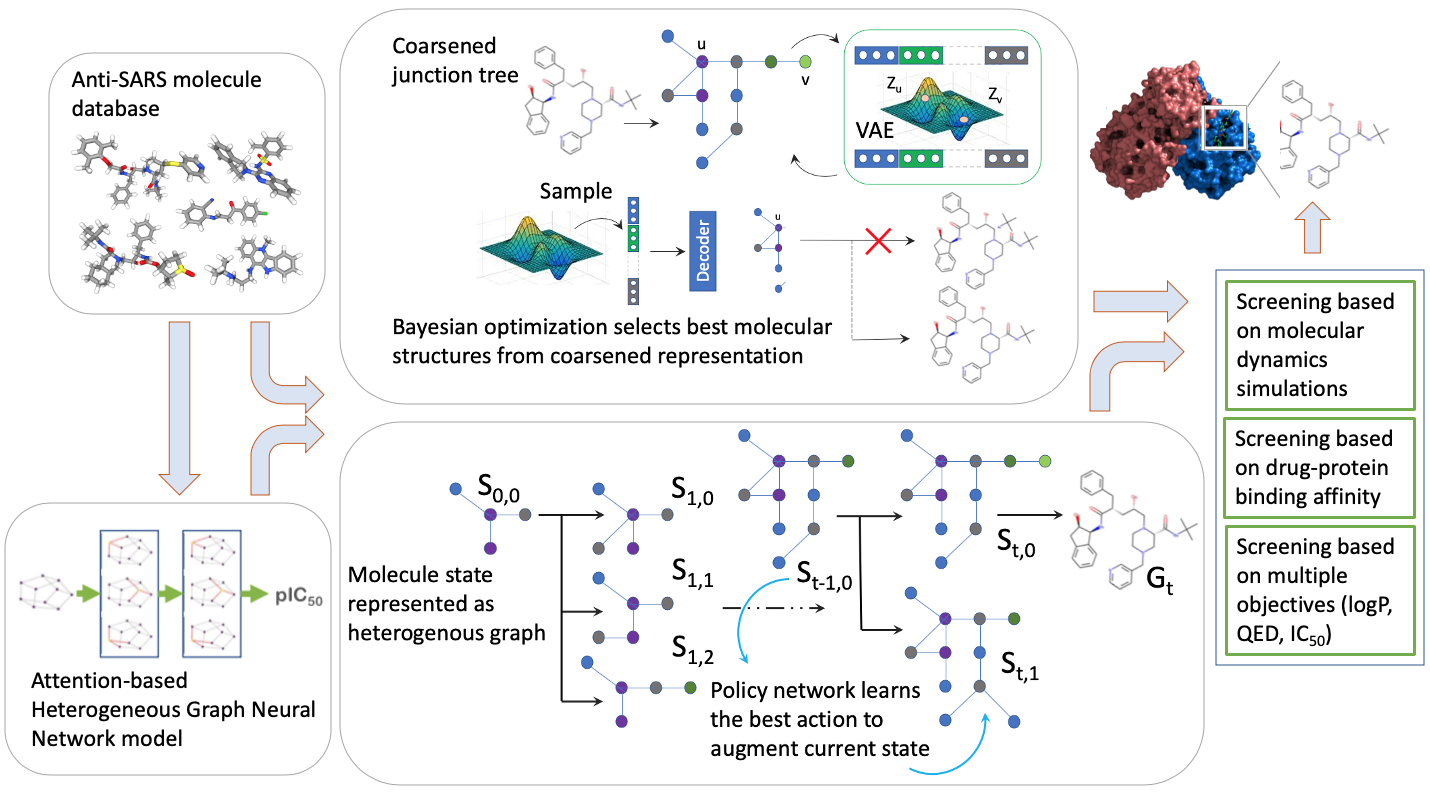}
  \caption{Depiction of workflow developed for this work. The anti-SARS database is used to train a MPNN to predict pIC\textsubscript{50} and the JT-VAE model. The trained MPNN is used as the scoring function in both JT-VAE (top row) and DQN-based molecular generation (bottom). Candidate molecules are screened by pIC\textsubscript{50} (>8) and validated by a Drug Target Binding Affinity classifier.} \label{fig:mol_design_architecture}
\end{figure*}

\section{Background and Related Work}

In response to the COVID-19 global pandemic, researchers have pushed to identify marketed drugs that can be repurposed for SARS-CoV-2 treatment \cite{COVID-options, COVID-repurpose-nature, COVID-repurpose, SummitDocking, CovidInteractionMap}. Born et al.\ amended their PaccMann RL approach, originally intended to generate anticancer drugs, to generate molecules with affinity to viral target proteins and controlled toxicity \cite{bornpaccmannrl}. Batra et al.\ applied an ML-based screening approach to find known compounds with binding affinity to either the isolated SARS-CoV-2 S-protein at its host receptor region or to the S-protein-human ACE2 interface complex \cite{Batra2020Covid}. Huang et al.\ developed a deep learning toolkit for drug repurposing, called DeepPurpose, with the goal of recommending candidates with high binding affinity to target amino acid sequences from known drugs \cite{huang2020deeppurpose}. These approaches screen large databases of known compounds, which have the potential to miss novel molecules with anti-SARS activity. Chenthamarakshan et al.\ developed the generative modeling framework CogMol to design drug candidates specific to a given target protein~\cite{IBM}.

%The discovery of novel drugs is often inefficient and costly due to the large initial search space. This search space can be reduced by generating novel molecules that are predicted to have desirable properties. Current molecular design algorithms are being adapted to the task of predicting potential anti-SARS drugs. 

\ian{``A variety of properties may be considered...}
Here we discuss which properties should be considered during lead optimization \cite{copeland2006drug}. LogP is a measure of lipophilicity, which provides an understanding of the behavior of a drug in the body. Compounds intended for oral administration should have a logP no greater than 5, according to Lipinski's Rule of 5 \cite{lipinski2004lead}. Further analysis has shown that logP values between 1 and 3 may be more appropriate considering the effect of logP on absorption, distribution, metabolism, elimination, and toxicology (ADMET) properties \cite{logP2010}. Though oral bioavailability is an important factor, a sole focus on logP has the potential to screen out otherwise useful compounds \cite{Beyond5}. QED has been proposed as a more holistic druglikeness metric \cite{QED}, from 0 (low) to 1 (high).  Druglikeness provides a general metric for ADMET properties, but does not indicate the activity or effectiveness of a drug towards a specific target. The half maximal inhibitory concentration (IC\textsubscript{50}), on the other hand, provides a quantitative measure of the potency of a compound to inhibit a specific biological process. IC\textsubscript{50} is obtained by measurement, and no universal ab initio method of computing its value exists. A number of methods have been developed to approximate IC\textsubscript{50}, many based on QSPR and recently some based on machine learning \cite{IC50-QSAR, IC50-QM, IC50-ML, IC50-LSTM}. Similarity to known drugs is also an important factor in drug discovery \cite{Similarity2014, Similarity2018}, as is the ability to synthesize the molecule, which can be estimated by the synthetic accessibility (SA) score [from 1 (easy) to 10 (difficult)] \cite{SAScore}.

%Gordon et al. generated a SARS-CoV-2 interaction map and identified 69 potential compounds for drug repurposing.\cite{CovidInteractionMap}

\section{Method}
\subsection{Surrogate Model for pIC\textsubscript{50} Prediction}
\label{sec:mpnn}
We trained a message-passing neural network (MPNN) \cite{scarselli2008graph, gilmer2017neural} to predict pIC\textsubscript{50} (the inverse log of IC\textsubscript{50}) for a given molecular structure.
Following the formalism of Gilmer et al.~\cite{gilmer2017neural}, our network is composed of message, update and readout operations (eqns.\ 1-3) and our choices
for these functions are based on networks developed by St. John et al. for polymer property prediction.~\cite{stjohn2019mpnnpolymer}

The original state of each atom ($h_v$) and bond ($\alpha_{vw}$) in our molecule ($G$) is a 256-length vector with values defined by an embedding table based on the atomic number and bond type (e.g., single, double, aromatic).
The states of these atoms are modified by eight successive ``message'' layers.
Each message layer uses a two-layer multi-layer perceptron (MLP) with sigmoid activations to compute a message that uses the state of an 
atom ($h_v$), the state of the neighboring atom ($h_w$) and the bond
which joins them ($\alpha_{vw}$). 
The atom and bond states are updated according to the following equations:

\begin{equation}
m^{t+1}_v = \sum_{w \in Neighbors(v)} M_t(h^t_v, h^t_w, \alpha^{t}_{vw})
\end{equation}
\begin{equation}
h^{t+1}_v = h^t_v + m^{t+1}_v
\end{equation}
\begin{equation}
\alpha^{t+1}_{vw} = \alpha^t_{vw} + M_t(h^t_v, h^t_w, \alpha^{t}_{vw})
\end{equation}

The atom states output from the last layer ($h^T_v$) are used to predict the $pIC_{50}$ of the molecule using a "readout" function ($R$).

\begin{equation}
\hat{y}=R(h^T_v |v \in G)
\label{eqn:ic50_aggr}
\end{equation}

We use several variants of the readout function in our study.
We tested both ``molecular fingerprints,'' where the states of each node are combined \textit{before} using a multi-layer perceptron (MLP) to reduce to compute pIC\textsubscript{50}, and an ``atomic contribution,'' where we combine the nodes \textit{after} MLP to compute a per-node contribution to pIC\textsubscript{50}.
We experimented with the use of five different functions to reduce the atomic state/contributions to a single value for each graph: summation, mean, maximum, softmax, and attention.
The attention functions are created by learning an attention map by passing the node states through a MPNN.
We tested all combinations of ``molecular fingerprint'' vs.\ ``atomic contribution'' and the five readout functions, for a total of 10 networks, training each on network the same 90\% of our  pIC\textsubscript{50} dataset and comparing its performance on the withheld 10\% of the data.
We used an MPNN that uses attention maps to reduce contributions from each atom to a 
single pIC\textsubscript{50} of a molecule in all subsequent experiments.

%Full details of our implementation are included on our GitHub page (URL excluded to respect double-blind reviewing).%(\url{https://github.com/exalearn/covid-drug-design}).

% \hfill \break
% \end{comment}

\subsection{Junction Tree Variational Autoencoder}

We use a junction-tree (JT) based variational autoencoder (VAE) method \cite{JTVAE} for generating molecules with high proximity to anti-SARS drug molecules.  This model generates novel molecular graphs by laying a tree-structured scaffold over substructures in the molecule, which are then combined into a valid molecule using a MPNN. The JT-VAE model allows for the incremental expansion of a molecular graph through the addition of subgraphs, or ``vocabulary" of valid components (denoted $V_D$), derived from the training set (Figure \ref{fig:mol_design_architecture}). The subgraphs are used to encode a molecule into its vector representation and decode latent vectors into valid molecular graphs. The use of subgraphs, rather than building a molecule atom-by-atom, maintains chemical validity at each step, while also incorporating functional groups common to the training set. Chemical graphs generated from the vocabulary will be structurally similar to those in the training set, which is a benefit when attempting to design drugs with similar properties to those in the training set. 

Given a molecular graph $G = (V, E)$, JT-VAE coarsens $G$ into a junction tree data structure $\mathcal{T} = (V_T, E_T)$ such that each node $v \in V_T$ in $\mathcal{T}$ represents a subgraph of $G$.  The subgraphs can be coarsened by using either automated JT construction algorithms from the graphical model literature \cite{lauritzen1988local} or a vocabulary based mapping approach that reduces to important building blocks of chemical structures.  
%The vocabulary generated from our anti-SARS training set contained 331 entries, with 15 types of atoms (H, C, N, O, P, S, B, Si, F, Cl, Br, I, Mn, Cu, and Co). The composition of the vocabulary is as follows: 102 have aromatic moieties, 224 have at least 1 hydrogen bond acceptor, 117 have at least 1 hydrogen bond donor, and no units contain rotatable bonds. The smallest unit contains only 1 atom, while the largest contains 24 atoms, with the mean being 6 atoms. The molecular weight ranges from 13.84 - 356.65, with a mean of 88.04 Da. The molecular polar surface area ranges from 0 - 57.36, with a mean of 20.45.

Next, the JT-VAE method uses a message-passing network (as described in section \ref{sec:mpnn}) to encode $\mathcal{T}$ (Figure \ref{fig:mol_design_architecture}) into a vector representation $\mathbf{Z}_{\mathcal{T}}$.  We refer the reader to \cite{JTVAE} for specific implementations of the MPNN-based encoders for the input graph and the junction tree.  The decoder component of the VAE learns to generate the same junction tree structure from $\mathbf{Z}_{\mathcal{T}}$ by maximizing the likelihood function $p(\mathcal{T}|\mathbf{Z}_{\mathcal{T}})$.

Once the VAE model is trained, the next phase involves a two-step process for generating a molecular graph structure that is optimized for a target property.  The first step involves drawing a sample from the latent space learnt by the VAE and transforming the sampled vector representation into a corresponding junction tree structure. The second step pursues a Bayesian optimization (BO) strategy to map the junction tree to a molecular graph that maximizes the target properties. Using the JT-VAE trained on our SARS-related database, we perform Bayesian optimization (BO), using the method of Kusner et al.~\cite{GrammarBO} to produce novel molecules with target properties described in section 3.4.

% Jin et al. expanded upon their original JT-VAE concept by replacing the vocabulary with rationales, molecular subgraphs chosen to have a certain predicted property score.\cite{JTVAE-multi} However, as we are attempting to improve molecular generation through tailoring the scoring function, we do not incorporate this added complexity here and instead employ the original JT-VAE algorithm.

\subsection{Deep Reinforcement Learning}

We follow the Q-learning approach of Zhou et al.\ for our deep reinforcement learning approach \cite{MolDQN}. We consider tasks in which an agent interacts with an environment $\mathcal{E}$ represented as a molecular graph.
The agent starts with a null graph.  
At each time-step the agent selects an action $a_t$ from an action space $\mathcal{A}$ that includes addition of single atoms, changing the type of bonds or removing bonds from the graph. 
The agent also receives a reward $\mathcal{R}$ at each time step depending on a property scoring function.  All the property scoring functions are described in section 3.4.

In this setting, we cast the molecule generation problem as a Markov Decision Process (MDP) \cite{DQN} to learn a policy network $\pi$ that determines the best sequence of actions that start with an initial molecular graph and transform it a larger graph with desirable properties in a step-by-step fashion (Figure \ref{fig:mol_design_architecture}).  At each step, we enumerate all possible actions and then select those which produce valid molecules (e.g., respect valency rules).  Next, we train a multi-layer perceptron (MLP) to predict to predict the \textsl{value} \cite{DQN} of a certain action by passing the Morgan fingerprints \cite{rogers2010extended} as input.  The MLP approximates the value of an action computed using the Bellman equation, where the score of a state and the maximum score of the subsequent state is multiplied by a decay factor.  As established with other Deep Q-Learning approaches, the addition of the value of the next state increases the value of moves which will lead to higher future rewards.

%Full details of our implementation and source code are included on our GitHub page (URL excluded to respect double-blind reviewing).%(\url{https://github.com/exalearn/covid-drug-design}).

\subsection{Scoring Functions}
\label{sec:Scoring Functions}

The scoring functions described in this section are used for both Bayesian optimization in the VAE based approach and reward computation for the deep reinforcement-learning based approach. Following Jin et al.~\cite{JTVAE}, we first compute a score that penalizes logP by the SA score (recall from section 2 that higher SA values are discouraged) and number of cycles with more than 6 atoms (eqn.~5). 
%which synthetic accessibility as well as the pharmacokinetic profile.\cite{WANG2019880} This scheme was applied by \cite{DDR1} for the development of an RL algorithm to generate DDR1 kinase inhibitors. 
Considering that QED is a more comprehensive heuristic than logP, we also use a similar scoring function composed of QED penalized by the SA score and number of long cycles (eqn.~6). We then examine the utility of a SARS-specific scoring function based on the pIC\textsubscript{50} predicted by our MPNN penalized by the SA score and number of long cycles. Finally, we examine a multi-objective scoring function that accommodates both pIC\textsubscript{50} and penalized QED. 

\begin{equation}
logP^P(m) = logP(m) - SA(m) - cycle(m)
\end{equation}
\begin{equation}
QED^P(m) = QED(m) - SA(m) - cycle(m)
\end{equation}
\begin{equation}
pIC_{50}(m) = R(graph(m)) \text{  (from eqn.~4)}%pIC_{50}(m)
\end{equation}
\begin{equation}
pIC_{50}+QED^P(m) = pIC_{50}(m) + QED(m) - SA(m) - cycle(m)
\end{equation}

\section{Experimental Analysis}
We conduct experiments to answer two questions: \textbf{Q1:} What are the best possible ways to generate molecules with targeted activity towards SARS-CoV-2?  \textbf{Q2:} How do we evaluate the novelty of our generated molecules? 
%We evaluate the generated molecules by the logP, SA score, QED, and pIC\textsubscript{50} to examine the effect of the different scoring functions based on these values. We then examine two methods for the prediction of protein binding to further validate our generated molecules as potential anti-SARS-CoV-2 candidates.

% \subsection{Experimental Setup}
\subsection{Dataset and Analysis}

We assembled a dataset with molecules active against various enzymatic assays filtered from experimental pharmacology data logged in CheMBL, BindingDB, and ToxCat \cite{bento2014chembl}. Details of this dataset are provided in section \ref{sec:Dataset Preparation}.  Following the workflow in Figure 1, we first train a pIC\textsubscript{50} model, and subsequently use it to train the JT-VAE and DQN models.  We begin with a discussion of the pIC\textsubscript{50} model given its key role in training the JT-VAE and DQN models.
% The database was filtered out with the IC\textsubscript{50} activity standard types and their potency. For a generative perspective, molecules with sizes larger than 1,000 Dalton were removed due to the limitation of the representation of large molecules in cheminformatics. 
% \ref{Gaulton, A. et al. ChEMBL: a large-scale bioactivity database for drug discovery. Nucleic
% Acids Res. 40, D1100–7 (2012).

% CHEMBL database release 25.
% http://ftp.ebi.ac.uk/pub/databases/chembl/ChEMBLdb/releases/chembl_25 (2019)
% doi:10.6019/CHEMBL.database.25.
% }
\subsection{Training pIC\textsubscript{50} surrogate model}

%\textbf{MPNN variants.} 
We tested  multiple variants of the pIC\textsubscript{50} prediction model.
We found that MPNNs that limited the computation of the pIC\textsubscript{50} to contributions from only a few
specific atoms in the molecule performed the best. As shown in Table~\ref{tab:mpnn_results}, the models which use max and softmax
functions for aggregating the atom-level representations to a molecule-level representation (equation \ref{eqn:ic50_aggr}) have higher $R^2$ scores than those which use summation or mean outputs.
% Attention-based models perform roughly in the middle.

\logan{Needs Neeraj/Sutanay's review}
The relative performance of different networks can be explained by the physical
mechanism behind the performance of anti-viral drugs. 
The presence or absence of a specific pattern in the molecular structure (e.g., functional groups, substituents) controls whether the molecule will bind with a certain portion
of a target virus protein. We hypothesize that the ``maximum'' function, in particular, models this ``all-or-nothing'' physics well.
The other atoms in the molecule play a role in determining whether the molecule will stay affixed at the target site. 
The reduced but non-negligible effect of the side groups could explain why the ``atomic''-contribution model, which only uses the contribution from a single atom,
performs less well than the whole-molecule fingerprint.
These trends give us confidence that the MPNNs are operating based on well-known physics,
but we would require comparison of which atoms are contributing to
the predicted pIC\textsubscript{50} to results from molecular binding simulations to determine 
if the networks are indeed interpretable.

% Jenna: commenting out for space
%We note that the model used for our experiments, the atomic contribution with attention, is the not the best-performing model. 
%We selected the atomic-attention model to explore whether attention can  be used to identify the key molecular substructure and plan to repeat with our best-performing model in future work.
%\logan{Maybe fix this for the revision?}

\subsection{Effect of the scoring function} \label{scoringsection}

\begin{figure}[htbp!] \centering
  \includegraphics[clip, trim=3cm 1cm 3cm 2cm, width=0.45\textwidth]{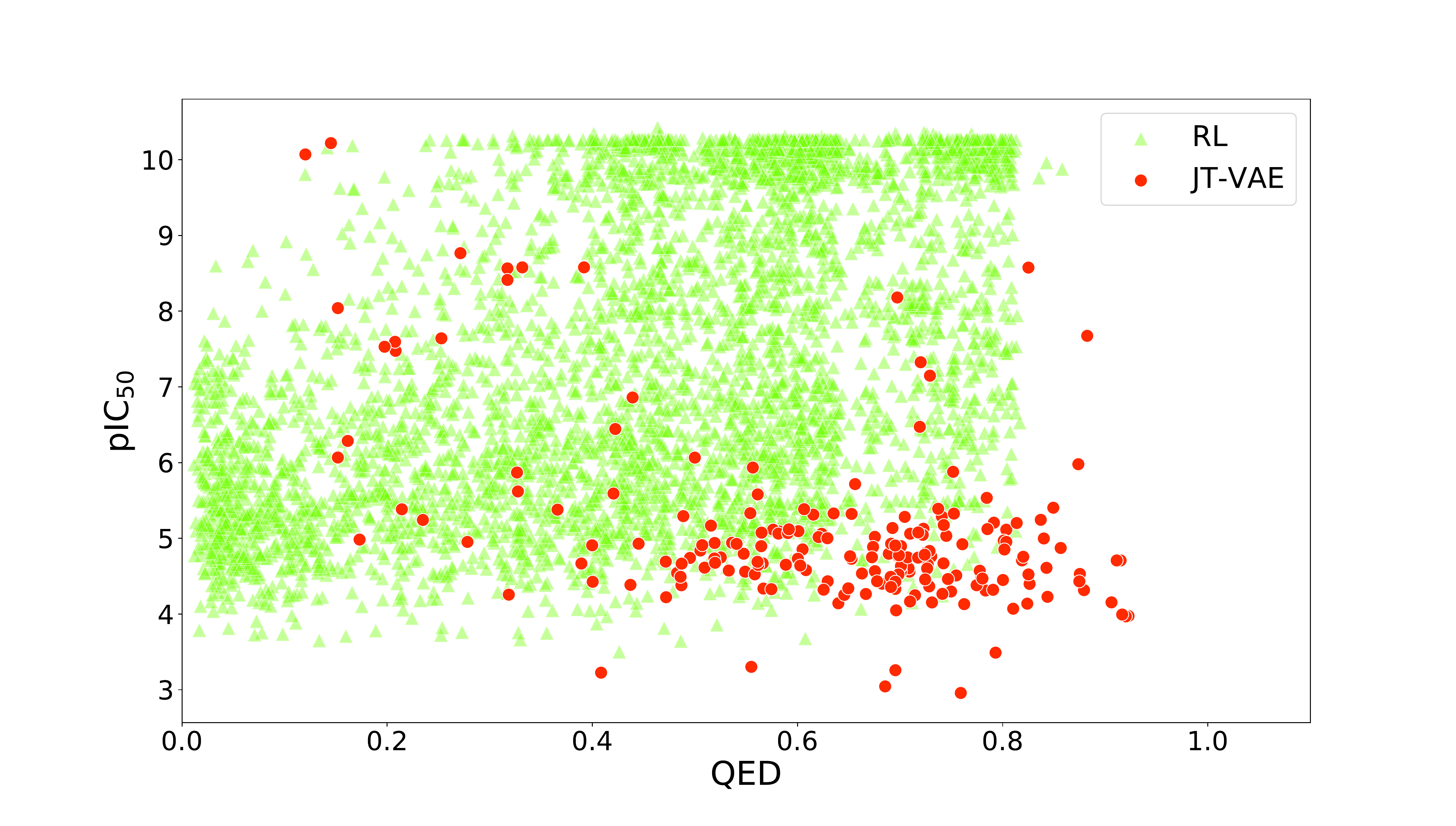}
%  \caption{QED vs pIC\textsubscript{50} plot comparing the molecules generated by RL and JT-VAE.}
\caption{QED versus pIC\textsubscript{50} for molecules generated from DQN-based RL and JT-VAE with all scoring functions. %The green triangles represent 4,455 unique molecules corresponding to the last step of each RL episode. The red circles represent 201 unique molecules generated from JT-VAE. 
}
\label{fig:pareto_plot}
\end{figure}

\textbf{Molecule generation setup using JT-VAE.} We trained JT-VAE for 8,300 iterations on the full database, with the following hyperparameters: hidden state dimension of 450, latent code dimension of 56, and graph message passing depth of 3. Analysis of the trained JT-VAE is given in Figure \ref{fig:DBvVAE}. To optimize towards the specified scoring functions, we trained a sparse Gaussian process (SGP) to predict a score given the latent representation learned by JT-VAE and then perform 10 iterations of batched BO (sampling = 50) based on the expected improvement. %This was done for each scoring function. %The results are given in our study of the scoring function in section \ref{scoringsection}.

% commented out for space
%For the JT-VAE, optimizing towards logP$^P$ and QED$^P$ led to good druglikeness, but low pIC\textsubscript{50}. Optimizing towards pIC\textsubscript{50} greatly increased the pIC\textsubscript{50} of the generated candidates, but at the expense of QED. Optimizing towards both pIC\textsubscript{50} and QED$^P$ improved the QED, while slightly decreasing the pIC\textsubscript{50}. A comparison of the properties of all molecules generated by the JT-VAE optimized with different scoring functions is given in Figure \ref{fig:JTVAEcomparison}.
\logan{I moved this from later in this section to make a focused section on comparing the performance.}

\begin{table*}
\caption{Properties of the three highest-scoring molecules generated by each model using the specified scoring function. %The pIC\textsubscript{50} value was estimated using our MPNN.\logan{I removed the score column as Jenna and I must have used different normalization schemes, making them incomparable. Also, I caught a bug in the QED program and have to re-run it. Will put in prelimary results, but will have to wait for the full model to finish}
}
  \resizebox{\textwidth}{!}{\begin{tabular}{lSSSSSSSSSSSSSSS}
    \toprule
    \multirow{2}{*}{Scoring function } &
      \multicolumn{3}{c}{pIC\textsubscript{50}} &
      \multicolumn{3}{c}{QED} &
      \multicolumn{3}{c}{logP} & 
      \multicolumn{3}{c}{SA Score}\\
       & {1st} & {2nd} & {3rd} & {1st} & {2nd} & {3rd} & {1st} & {2nd} & {3rd}& {1st} & {2nd} & {3rd}\\
      \midrule
    logP$^P$ (JT-VAE)	&	4.93	&	4.57	&	4.60	&	0.45	&	0.78	&	0.71	&	4.05	&	4.27	&	4.20	&	1.70	&	2.08	&	2.08	\\
    logP$^P$ (DQN)  &  6.10 &  8.17 &  4.98 &  0.04 &  0.07 &  0.11 &  13.86 &  12.64 &  12.52 & 3.59 & 2.97 & 2.93 \\
    \hline
    QED$^P$	 (JT-VAE)&	4.15	&	4.23	&	4.71	&	0.91	&	0.84	&	0.91	&	3.72	&	2.19	&	2.20	&	1.80	&	1.71	&	2.25	\\
    QED$^P$	(DQN) &  6.80 &  6.80 &  6.80 &  0.77 &  0.77 &  0.77 &  3.46 &  3.46 &  3.46 & 2.05 & 2.05 & 2.05	\\
    \hline
    pIC\textsubscript{50} (JT-VAE)	&	10.22	&	10.07	&	10.07	&	0.15	&	0.12	&	0.12	&	4.86	&	3.80	&	3.06	&	4.52	&	4.94	&	4.98	\\
    pIC\textsubscript{50} (DQN) &  10.57 &  10.56 &  10.39 &  0.09 &  0.11 &  0.41 &  1.51 &  2.03 & -0.44 & 6.90 & 6.71 & 5.57 \\
    \hline
    pIC\textsubscript{50}+QED$^P$ (JT-VAE)  &	8.58	&	5.98	&	8.18	&	0.83	&	0.87	&	0.70	&	4.02	&	3.37	&	3.82	&	1.93	&	1.74	&	1.50\\
    pIC\textsubscript{50}+QED$^P$ (DQN)   &  10.27 &  10.27 &  10.27 &  0.80 &  0.80 &  0.80 &  3.02 &  3.02 &  3.02 & 2.90 & 2.90 & 2.90 \\
    \bottomrule
      \end{tabular}}\label{tab:jtvae}
\end{table*}

\textbf{Molecule generation setup using DQN.}
Our DQN approach builds up molecules from a single atom to large molecules atom-by-atom and bond-by-bond.
Each ``episode'' starts with a blank slate and the RL agent is allowed up to 40 steps.
We update a model of the Q-function to predict the value of each move after each step in each 
episode and, as this model improves during training, we smoothly turn down the 
probability that we will choose a random move over the prediction of this model. 

% commenting out for space
The DQN finds tens of thousands of candidate molecules with high p\icfifty, as shown in Figure~\ref{fig:pareto_plot} and Table 1.
%, as shown in Figure~\ref{fig:rl_traject}. 
Using the multi-objective reward function led to fewer molecules with p\icfifty $ > 8$, but with increased druglikeness.
%but that some notion of druglikeness is required in the reward function to bias the search towards drug-like models. 
Approximately two-thirds of the molecules with p\icfifty$> 8$ found with the multi-objective reward also show QED$ > 0.5$. In contrast, only $5\%$ of the molecules in the p\icfifty-based search show QED > 0.5. The addition of QED reduces the total number of high-p\icfifty molecules found by 25\%, but increases the number of high-p\icfifty molecules found by over 10 times. Therefore, we recommend incorporating synthesizability and/or druglikeness into RL-driven searches for drug-like molecules.
\logan{Be consistent about ``scoring'' function and not reward function.}

\textbf{Comparing JT-VAE and DQN.}
Table \ref{tab:jtvae} shows the three highest-scoring molecules generated by the two generative models with each scoring function. 
The DQN models always outperform JT-VAE in finding a molecule with a superior value of the scoring function being optimized.
The performance disparity is particularly apparent when optimizing for logP:
the maximum logP from DQN is 12.6 compared to only 4.1 for JT-VAE.
%The computational expense for each method is not dramatically different as each require several hours on a V100 GPU to generate the candidate molecules. 
We attribute the difference in optimization performance to JT-VAE implicitly sampling from a distribution of drug-like molecules and DQN having no such
constraints.

The candidate molecules generated by JT-VAE have consistently better druglikeness 
and SA scores even when those values are not explicitly optimized for. 
When optimizing towards logP, the top-3 molecules generated by JT-VAE have moderate-to-high QEDs, while the top-3 from DQN are below 0.11.
We attribute the exceptionally large disparity in optimal logP and associated QED values between the two methods to the fact that drugs typically have
logP values between -0.4 and 5.6. 
The molecules from which JT-VAE was trained were all drug-like molecules,
which makes it improbable to sample molecules with logP values far outside that range.
Similarly, the molecules in the JT-VAE training set were experimentally synthesized.
The SA score for these molecules is low, which could explain why the optimized
molecules from the JT-VAE are also low even when this property was not optimized for.
The RL agent uses no information about the space of experimentally studied drug molecules during its training process and, accordingly, finds molecules far from it.

Overall, we find two different purposes for JT-VAE and RL-based molecular optimization.
JT-VAE implicitly uses the distribution of molecules in its training set to 
bias towards realistic molecules, albeit at the expense of finding \ian{fewer good?} better candidates.
The RL-based approach lacks such constraints and, for better or worse, 
can optimize without even implicitly regarding synthesizability or any
other characteristic not explicitly encoded in the scoring function.

%For the JT-VAE, when first optimizing towards logP$^P$, the top-3 molecules have logP values between 4.05 and 4.27 and low SA scores. The QED values are moderate and the pIC\textsubscript{50} is low (<5). Optimizing towards QED$^P$ increases the QED, maintains the low SA score, decreases logP and has no effect on pIC\textsubscript{50}. Notably, neither the logP$^P$ or QED$^P$  scoring function generated any molecules with high pIC\textsubscript{50} values. Optimizing towards pIC\textsubscript{50} greatly increased the pIC\textsubscript{50} of the top-3 generated molecules to above 10. The QED of these molecules are very low, the logP remains moderate, and the SA score is increased but still under 5. Using the pIC\textsubscript{50}+QED$^P$ scoring function increased the QED and lowered the SA score, but at the expense of the pIC\textsubscript{50}; no effect was had on logP. A comparison of the properties of all molecules generated by the JT-VAE optimized with different scoring functions can be seen in Figure \ref{fig:JTVAEcomparison}.

\subsection{Qualitative analysis from a drug discovery perspective}

\begin{figure}[htbp!] \centering
  \includegraphics[clip, trim=6.3cm 4cm 5cm 4cm, width=0.48\textwidth]{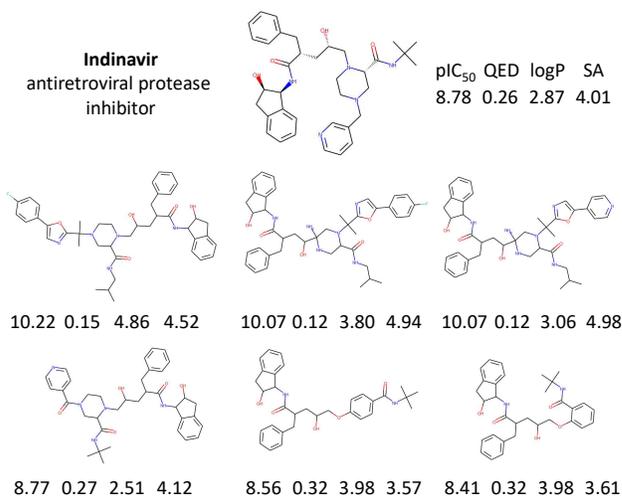}
  \caption{The top-6 molecules generated by the JT-VAE method optimized using the pIC\textsubscript{50} scoring function. %The pIC\textsubscript{50}, logP, QED, and SA score, in that order, are listed below each molecule. The pIC\textsubscript{50} value was estimated using our MPNN.
  }
  \label{fig:topmolecules}
\end{figure}

We observed an interesting structural trend in the molecules generated by JT-VAE when using pIC\textsubscript{50} as the scoring function. Figure \ref{fig:topmolecules} shows the structures of both the molecules with pIC\textsubscript{50} > 8 and the anti-HIV drug Indinavir, an antiretroviral protease inhibitor. A common backbone is shared between Indinavir and the top-6 predictions. The Tanimoto similarity scores \cite{Tanimoto2015} of these six generated molecules against Indinavir range from 0.65 to 0.91. Indinavir has been proposed as a drug to treat SARS-CoV-2 due to favorable docking to the coronavirus 3-chymotrypsin-like protease (3CL-protease), a promising drug target for combating coronavirus infections \cite{Indinavirdocking, 3CLforCOVID,harrison2020coronavirus}.
%indicating that the generated molecules have very high structural similarity to this candidate Alkaloids have also been indicated to have activity against 3CL-protease, and all generated molecules contain alkaloid-like funcationalities.\cite{3CLforCOVID} 
Notably, three of the generated molecules have a higher predicted pIC\textsubscript{50} than does Indinavir. %Notably, the top-4th molecule is only structurally different from Indinavir by a single carbonyl and has an equivalent pIC\textsubscript{50} score, as well as similar QED, logP, and SA score, which may indicate that this candidate would show a similar biological effect. 

%In total, we generated 222 unique molecules using JT-VAE (37\% unique rate). % In addition, connection between RL and JT-VAE
%The unique compounds generated by RL were analysed by computing their molecular weights. Out of 4,455 compounds, 4,283 were found to have molecular weights larger than 200 Dalton.  This indicates that the RL-generated molecules are large enough to be considered as drug-like compounds or possible precursors for constructing larger molecules. It is important to highlight that the sampling capability of RL is unrestricted, unlike the a priori training requirement of the JT-VAE. Hence, more chemically intuitive filtering rules should be enforced in RL to constrain the search space.%  (e.g. space of commercially purchasable compounds, network terminate early when there is no valence atoms to add additional functionality etc).  

% \subsection{DTBA model}
%\textbf{Evaluation based on Drug Target Binding Affinity (DTBA).}
Experimental confirmation of Drug-Target-Interaction (DTI) is challenging and time-consuming~\cite{thafar2019comparison}. \emph{In silico} Drug-Target Binding Affinity (DTBA) methods offer an alternative to evaluate DTI~\cite{he2017simboost}.  We employ a ML-based DTBA model to validate the interaction of molecules generated by  JT-VAE against a target SARS-CoV2-L protease~\cite{chen2020prediction}. We trained a DBTA binary classification model using extended connectivity fingerprint~\cite{rogers2010extended} encoding for the drug molecule and the target protease sequence encoding using a Convolution Neural Network (CNN) as implemented in the DeepPurpose toolkit~\cite{huang2020deeppurpose}. The default hyperparameters provided in the DeepPurpose toolkit were found to be sufficient. The DBTA model classified four of the top 11 molecules (including the top two in Figure \ref{fig:topmolecules}) with probability $>$ 0.5 to have interaction with SARS-CoV2-3CL protease.

\section{Conclusions}
We compared two graph generative models, JT-VAE and DQN, for the task of discovering potential small-molecule candidates with activity against SARS-CoV-2. DQN always outperformed JT-VAE in finding a molecule with a superior value of the scoring function being optimized. However, JT-VAE generated molecules that were more structurally similar to those in the database due to substructure representation, which produced a lower SA score and logP < 5. JT-VAE tended to produce what looked to be drug molecules, while DQN produced precursor-like candidates with optimized properties, which could be used as starting structures to add additional substituents aimed at the specific target.

% commented out for space
%Bayesian optimization of a JT-VAE trained on an anti-SARS dataset using pIC\textsubscript{50} in the scoring function produces candidates targeted at anti-SARS activity. The top-6 molecules generated using pIC\textsubscript{50} as the scoring function have similar structures to the anti-HIV drug Indinavir, and the top-2 were predicted to have high affinity with SARS-CoV2-3CL protease.

\bibliographystyle{ACM-Reference-Format}
\bibliography{citations}


\begin{thebibliography}{52}

%%% the Web address.

\ifx \showCODEN    \undefined \def \showCODEN     #1{\unskip}     \fi
\ifx \showDOI      \undefined \def \showDOI       #1{#1}\fi
\ifx \showISBNx    \undefined \def \showISBNx     #1{\unskip}     \fi
\ifx \showISBNxiii \undefined \def \showISBNxiii  #1{\unskip}     \fi
\ifx \showISSN     \undefined \def \showISSN      #1{\unskip}     \fi
\ifx \showLCCN     \undefined \def \showLCCN      #1{\unskip}     \fi
\ifx \shownote     \undefined \def \shownote      #1{#1}          \fi
\ifx \showarticletitle \undefined \def \showarticletitle #1{#1}   \fi
\ifx \showURL      \undefined \def \showURL       {\relax}        \fi
% invisible to TeX
\providecommand\bibfield[2]{#2}
\providecommand\bibinfo[2]{#2}
\providecommand\natexlab[1]{#1}
\providecommand\showeprint[2][]{arXiv:#2}

\bibitem[\protect\citeauthoryear{Armutlu, Ozdemir, Uney-Yuksektepe, Kavakli,
  and Turkay}{Armutlu et~al\mbox{.}}{2008}]%
        {IC50-ML}
\bibfield{author}{\bibinfo{person}{P. Armutlu}, \bibinfo{person}{M.~E.
  Ozdemir}, \bibinfo{person}{F. Uney-Yuksektepe}, \bibinfo{person}{I.~H.
  Kavakli}, {and} \bibinfo{person}{M. Turkay}.}
  \bibinfo{year}{2008}\natexlab{}.
\newblock \showarticletitle{Classification of drug molecules considering their
  IC50 values using mixed-integer linear programming based hyper-boxes method}.
\newblock \bibinfo{journal}{\emph{BMC Bioinformatics}}  \bibinfo{volume}{9}
  (\bibinfo{year}{2008}), \bibinfo{pages}{411}.
\newblock
\showISSN{1471-2105}
\urldef\tempurl%
\url{https://doi.org/10.1186/1471-2105-9-411}
\showDOI{\tempurl}


\bibitem[\protect\citeauthoryear{Avdeef, Box, Comer, Hibbert, and Tam}{Avdeef
  et~al\mbox{.}}{1998}]%
        {avdeef1998ph}
\bibfield{author}{\bibinfo{person}{A Avdeef}, \bibinfo{person}{KJ Box},
  \bibinfo{person}{JEA Comer}, \bibinfo{person}{C Hibbert}, {and}
  \bibinfo{person}{KY Tam}.} \bibinfo{year}{1998}\natexlab{}.
\newblock \showarticletitle{pH-Metric logP 10. Determination of liposomal
  membrane-water partition coefficients of lonizable drugs}.
\newblock \bibinfo{journal}{\emph{Pharmaceutical Research}}
  \bibinfo{volume}{15}, \bibinfo{number}{2} (\bibinfo{year}{1998}),
  \bibinfo{pages}{209--215}.
\newblock


\bibitem[\protect\citeauthoryear{Bag and Ghorai}{Bag and Ghorai}{2016}]%
        {IC50-QM}
\bibfield{author}{\bibinfo{person}{A. Bag} {and} \bibinfo{person}{P.~K.
  Ghorai}.} \bibinfo{year}{2016}\natexlab{}.
\newblock \showarticletitle{Development of Quantum Chemical Method to Calculate
  Half Maximal Inhibitory Concentration (IC50)}.
\newblock \bibinfo{journal}{\emph{Mol Inform}} \bibinfo{volume}{35},
  \bibinfo{number}{5} (\bibinfo{year}{2016}), \bibinfo{pages}{199--206}.
\newblock
\showISSN{1868-1743}
\urldef\tempurl%
\url{https://doi.org/10.1002/minf.201501004}
\showDOI{\tempurl}


\bibitem[\protect\citeauthoryear{Bajusz, R{\'a}cz, and H{\'e}berger}{Bajusz
  et~al\mbox{.}}{2015}]%
        {Tanimoto2015}
\bibfield{author}{\bibinfo{person}{D{\'a}vid Bajusz}, \bibinfo{person}{Anita
  R{\'a}cz}, {and} \bibinfo{person}{K{\'a}roly H{\'e}berger}.}
  \bibinfo{year}{2015}\natexlab{}.
\newblock \showarticletitle{Why is Tanimoto index an appropriate choice for
  fingerprint-based similarity calculations?}
\newblock \bibinfo{journal}{\emph{Journal of cheminformatics}}
  \bibinfo{volume}{7}, \bibinfo{number}{1} (\bibinfo{year}{2015}),
  \bibinfo{pages}{20}.
\newblock


\bibitem[\protect\citeauthoryear{{Batra}, {Chan}, {Kamath}, {Ramprasad},
  {Cherukara}, and {Sankaranarayanan}}{{Batra} et~al\mbox{.}}{2020}]%
        {Batra2020Covid}
\bibfield{author}{\bibinfo{person}{Rohit {Batra}}, \bibinfo{person}{Henry
  {Chan}}, \bibinfo{person}{Ganesh {Kamath}}, \bibinfo{person}{Rampi
  {Ramprasad}}, \bibinfo{person}{Mathew~J. {Cherukara}}, {and}
  \bibinfo{person}{Subramanian {Sankaranarayanan}}.}
  \bibinfo{year}{2020}\natexlab{}.
\newblock \showarticletitle{{Screening of Therapeutic Agents for COVID-19 using
  Machine Learning and Ensemble Docking Simulations}}.
\newblock \bibinfo{journal}{\emph{arXiv e-prints}}, Article
  \bibinfo{articleno}{arXiv:2004.03766} (\bibinfo{date}{April}
  \bibinfo{year}{2020}), \bibinfo{numpages}{arXiv:2004.03766}~pages.
\newblock
\showeprint[arxiv]{q-bio.BM/2004.03766}


\bibitem[\protect\citeauthoryear{Bento, Gaulton, Hersey, Bellis, Chambers,
  Davies, Kr{\"u}ger, Light, Mak, McGlinchey, et~al\mbox{.}}{Bento
  et~al\mbox{.}}{2014}]%
        {bento2014chembl}
\bibfield{author}{\bibinfo{person}{A~Patr{\'\i}cia Bento},
  \bibinfo{person}{Anna Gaulton}, \bibinfo{person}{Anne Hersey},
  \bibinfo{person}{Louisa~J Bellis}, \bibinfo{person}{Jon Chambers},
  \bibinfo{person}{Mark Davies}, \bibinfo{person}{Felix~A Kr{\"u}ger},
  \bibinfo{person}{Yvonne Light}, \bibinfo{person}{Lora Mak},
  \bibinfo{person}{Shaun McGlinchey}, {et~al\mbox{.}}}
  \bibinfo{year}{2014}\natexlab{}.
\newblock \showarticletitle{The ChEMBL bioactivity database: an update}.
\newblock \bibinfo{journal}{\emph{Nucleic Acids Research}}
  \bibinfo{volume}{42}, \bibinfo{number}{D1} (\bibinfo{year}{2014}),
  \bibinfo{pages}{D1083--D1090}.
\newblock


\bibitem[\protect\citeauthoryear{Bickerton, Paolini, Besnard, Muresan, and
  Hopkins}{Bickerton et~al\mbox{.}}{2012a}]%
        {bickerton2012quantifying}
\bibfield{author}{\bibinfo{person}{G~Richard Bickerton},
  \bibinfo{person}{Gaia~V Paolini}, \bibinfo{person}{J{\'e}r{\'e}my Besnard},
  \bibinfo{person}{Sorel Muresan}, {and} \bibinfo{person}{Andrew~L Hopkins}.}
  \bibinfo{year}{2012}\natexlab{a}.
\newblock \showarticletitle{Quantifying the chemical beauty of drugs}.
\newblock \bibinfo{journal}{\emph{Nature Chemistry}} \bibinfo{volume}{4},
  \bibinfo{number}{2} (\bibinfo{year}{2012}), \bibinfo{pages}{90}.
\newblock


\bibitem[\protect\citeauthoryear{Bickerton, Paolini, Besnard, Muresan, and
  Hopkins}{Bickerton et~al\mbox{.}}{2012b}]%
        {QED}
\bibfield{author}{\bibinfo{person}{G.~Richard Bickerton},
  \bibinfo{person}{Gaia~V. Paolini}, \bibinfo{person}{Jérémy Besnard},
  \bibinfo{person}{Sorel Muresan}, {and} \bibinfo{person}{Andrew~L. Hopkins}.}
  \bibinfo{year}{2012}\natexlab{b}.
\newblock \showarticletitle{Quantifying the chemical beauty of drugs}.
\newblock \bibinfo{journal}{\emph{Nature chemistry}} \bibinfo{volume}{4},
  \bibinfo{number}{2} (\bibinfo{year}{2012}), \bibinfo{pages}{90--98}.
\newblock
\showISSN{1755-4349 1755-4330}
\urldef\tempurl%
\url{https://doi.org/10.1038/nchem.1243}
\showDOI{\tempurl}


\bibitem[\protect\citeauthoryear{Bjerrum}{Bjerrum}{2017}]%
        {IC50-LSTM}
\bibfield{author}{\bibinfo{person}{Esben~Jannik Bjerrum}.}
  \bibinfo{year}{2017}\natexlab{}.
\newblock \showarticletitle{{SMILES} Enumeration as Data Augmentation for
  Neural Network Modeling of Molecules}.
\newblock \bibinfo{journal}{\emph{CoRR}}  \bibinfo{volume}{abs/1703.07076}
  (\bibinfo{year}{2017}).
\newblock
\showeprint[arxiv]{1703.07076}
\urldef\tempurl%
\url{http://arxiv.org/abs/1703.07076}
\showURL{%
\tempurl}


\bibitem[\protect\citeauthoryear{Born, Manica, Cadow, Markert, Mill,
  Filipavicius, and Mart{\'\i}nez}{Born et~al\mbox{.}}{[n.d.]}]%
        {bornpaccmannrl}
\bibfield{author}{\bibinfo{person}{Jannis Born}, \bibinfo{person}{Matteo
  Manica}, \bibinfo{person}{Joris Cadow}, \bibinfo{person}{Greta Markert},
  \bibinfo{person}{Nil~Adell Mill}, \bibinfo{person}{Modestas Filipavicius},
  {and} \bibinfo{person}{Mar{\'\i}a~Rodr{\'\i}guez Mart{\'\i}nez}.}
  \bibinfo{year}{[n.d.]}\natexlab{}.
\newblock \showarticletitle{PaccMannRL on SARS-CoV-2: Designing antiviral
  candidates with conditional generative models}.
\newblock  (\bibinfo{year}{[n.\,d.]}).
\newblock


\bibitem[\protect\citeauthoryear{Chang}{Chang}{2020}]%
        {Indinavirdocking}
\bibfield{author}{\bibinfo{person}{Y.~et~al. Chang}.}
  \bibinfo{year}{2020}\natexlab{}.
\newblock \showarticletitle{Potential Therapeutic Agents for COVID-19 Based on
  the Analysis of Protease and RNA Polymerase Docking.}
\newblock \bibinfo{journal}{\emph{Preprints}} (\bibinfo{year}{2020}),
  \bibinfo{pages}{2020020242}.
\newblock
\urldef\tempurl%
\url{https://doi.org/10.20944/preprints202002.0242.v1}
\showDOI{\tempurl}


\bibitem[\protect\citeauthoryear{Chen, Yiu, and Wong}{Chen
  et~al\mbox{.}}{2020}]%
        {chen2020prediction}
\bibfield{author}{\bibinfo{person}{Yu~Wai Chen},
  \bibinfo{person}{Chin-Pang~Bennu Yiu}, {and} \bibinfo{person}{Kwok-Yin
  Wong}.} \bibinfo{year}{2020}\natexlab{}.
\newblock \showarticletitle{Prediction of the SARS-CoV-2 (2019-nCoV) 3C-like
  protease (3CL pro) structure: virtual screening reveals velpatasvir,
  ledipasvir, and other drug repurposing candidates}.
\newblock \bibinfo{journal}{\emph{F1000Research}}  \bibinfo{volume}{9}
  (\bibinfo{year}{2020}).
\newblock


\bibitem[\protect\citeauthoryear{Chenthamarakshan, Das, Padhi, Strobelt, Lim,
  Hoover, Hoffman, and Mojsilovic}{Chenthamarakshan et~al\mbox{.}}{2020}]%
        {IBM}
\bibfield{author}{\bibinfo{person}{Vijil Chenthamarakshan},
  \bibinfo{person}{Payel Das}, \bibinfo{person}{Inkit Padhi},
  \bibinfo{person}{Hendrik Strobelt}, \bibinfo{person}{Kar~Wai Lim},
  \bibinfo{person}{Ben Hoover}, \bibinfo{person}{Samuel~C Hoffman}, {and}
  \bibinfo{person}{Aleksandra Mojsilovic}.} \bibinfo{year}{2020}\natexlab{}.
\newblock \showarticletitle{Target-specific and selective drug design for
  covid-19 using deep generative models}.
\newblock \bibinfo{journal}{\emph{arXiv preprint arXiv:2004.01215}}
  (\bibinfo{year}{2020}).
\newblock


\bibitem[\protect\citeauthoryear{Copeland, Pompliano, and Meek}{Copeland
  et~al\mbox{.}}{2006}]%
        {copeland2006drug}
\bibfield{author}{\bibinfo{person}{Robert~A Copeland}, \bibinfo{person}{David~L
  Pompliano}, {and} \bibinfo{person}{Thomas~D Meek}.}
  \bibinfo{year}{2006}\natexlab{}.
\newblock \showarticletitle{Drug--target residence time and its implications
  for lead optimization}.
\newblock \bibinfo{journal}{\emph{Nature Reviews Drug Discovery}}
  \bibinfo{volume}{5}, \bibinfo{number}{9} (\bibinfo{year}{2006}),
  \bibinfo{pages}{730--739}.
\newblock


\bibitem[\protect\citeauthoryear{de~Wit, van Doremalen, Falzarano, and
  Munster}{de~Wit et~al\mbox{.}}{2016}]%
        {de2016sars}
\bibfield{author}{\bibinfo{person}{Emmie de Wit}, \bibinfo{person}{Neeltje van
  Doremalen}, \bibinfo{person}{Darryl Falzarano}, {and}
  \bibinfo{person}{Vincent~J Munster}.} \bibinfo{year}{2016}\natexlab{}.
\newblock \showarticletitle{SARS and MERS: recent insights into emerging
  coronaviruses}.
\newblock \bibinfo{journal}{\emph{Nature Reviews Microbiology}}
  \bibinfo{volume}{14}, \bibinfo{number}{8} (\bibinfo{year}{2016}),
  \bibinfo{pages}{523}.
\newblock


\bibitem[\protect\citeauthoryear{Dror, Dirks, Grossman, Xu, and Shaw}{Dror
  et~al\mbox{.}}{2012}]%
        {dror2012biomolecular}
\bibfield{author}{\bibinfo{person}{Ron~O Dror}, \bibinfo{person}{Robert~M
  Dirks}, \bibinfo{person}{JP Grossman}, \bibinfo{person}{Huafeng Xu}, {and}
  \bibinfo{person}{David~E Shaw}.} \bibinfo{year}{2012}\natexlab{}.
\newblock \showarticletitle{Biomolecular simulation: a computational microscope
  for molecular biology}.
\newblock \bibinfo{journal}{\emph{Annual Review of Biophysics}}
  \bibinfo{volume}{41} (\bibinfo{year}{2012}), \bibinfo{pages}{429--452}.
\newblock


\bibitem[\protect\citeauthoryear{Ertl and Schuffenhauer}{Ertl and
  Schuffenhauer}{2009}]%
        {SAScore}
\bibfield{author}{\bibinfo{person}{Peter Ertl} {and} \bibinfo{person}{Ansgar
  Schuffenhauer}.} \bibinfo{year}{2009}\natexlab{}.
\newblock \showarticletitle{Estimation of synthetic accessibility score of
  drug-like molecules based on molecular complexity and fragment
  contributions}.
\newblock \bibinfo{journal}{\emph{Journal of Cheminformatics}}
  \bibinfo{volume}{1}, \bibinfo{number}{1} (\bibinfo{year}{2009}),
  \bibinfo{pages}{8}.
\newblock
\showISSN{1758-2946}
\urldef\tempurl%
\url{https://doi.org/10.1186/1758-2946-1-8}
\showDOI{\tempurl}


\bibitem[\protect\citeauthoryear{Fan, Wang, Liu, An, Liu, He, Song, and
  Tong}{Fan et~al\mbox{.}}{2020}]%
        {COVID-repurpose}
\bibfield{author}{\bibinfo{person}{Hua-Hao Fan}, \bibinfo{person}{Li-Qin Wang},
  \bibinfo{person}{Wen-Li Liu}, \bibinfo{person}{Xiao-Ping An},
  \bibinfo{person}{Zhen-Dong Liu}, \bibinfo{person}{Xiao-Qi He},
  \bibinfo{person}{Li-Hua Song}, {and} \bibinfo{person}{Yi-Gang Tong}.}
  \bibinfo{year}{2020}\natexlab{}.
\newblock \showarticletitle{Repurposing of clinically approved drugs for
  treatment of coronavirus disease 2019 in a 2019-novel coronavirus-related
  coronavirus model}.
\newblock \bibinfo{journal}{\emph{Chinese medical journal}}
  \bibinfo{volume}{133}, \bibinfo{number}{9} (\bibinfo{year}{2020}),
  \bibinfo{pages}{1051--1056}.
\newblock
\showISSN{2542-5641 0366-6999}
\urldef\tempurl%
\url{https://doi.org/10.1097/CM9.0000000000000797}
\showDOI{\tempurl}


\bibitem[\protect\citeauthoryear{Gilmer, Schoenholz, Riley, Vinyals, and
  Dahl}{Gilmer et~al\mbox{.}}{2017}]%
        {gilmer2017neural}
\bibfield{author}{\bibinfo{person}{Justin Gilmer}, \bibinfo{person}{Samuel~S
  Schoenholz}, \bibinfo{person}{Patrick~F Riley}, \bibinfo{person}{Oriol
  Vinyals}, {and} \bibinfo{person}{George~E Dahl}.}
  \bibinfo{year}{2017}\natexlab{}.
\newblock \showarticletitle{Neural message passing for quantum chemistry}. In
  \bibinfo{booktitle}{\emph{34th International Conference on Machine
  Learning-Volume 70}}. JMLR. org, \bibinfo{pages}{1263--1272}.
\newblock


\bibitem[\protect\citeauthoryear{Gordon, Jang, Bouhaddou, Xu, Obernier, White,
  O’Meara, Rezelj, Guo, Swaney, Tummino, Huettenhain, Kaake, Richards,
  Tutuncuoglu, Foussard, Batra, Haas, Modak, Kim, Haas, Polacco, Braberg,
  Fabius, Eckhardt, Soucheray, Bennett, Cakir, McGregor, Li, Meyer, Roesch,
  Vallet, Mac~Kain, Miorin, Moreno, Naing, Zhou, Peng, Shi, Zhang, Shen, Kirby,
  Melnyk, Chorba, Lou, Dai, Barrio-Hernandez, Memon, Hernandez-Armenta, Lyu,
  Mathy, Perica, Pilla, Ganesan, Saltzberg, Rakesh, Liu, Rosenthal, Calviello,
  Venkataramanan, Liboy-Lugo, Lin, Huang, Liu, Wankowicz, Bohn, Safari, Ugur,
  Koh, Savar, Tran, Shengjuler, Fletcher, O’Neal, Cai, Chang, Broadhurst,
  Klippsten, Sharp, Wenzell, Kuzuoglu, Wang, Trenker, Young, Cavero, Hiatt,
  Roth, Rathore, Subramanian, Noack, Hubert, Stroud, Frankel, Rosenberg, Verba,
  Agard, Ott, Emerman, Jura, et~al\mbox{.}}{Gordon et~al\mbox{.}}{2020}]%
        {CovidInteractionMap}
\bibfield{author}{\bibinfo{person}{David~E. Gordon},
  \bibinfo{person}{Gwendolyn~M. Jang}, \bibinfo{person}{Mehdi Bouhaddou},
  \bibinfo{person}{Jiewei Xu}, \bibinfo{person}{Kirsten Obernier},
  \bibinfo{person}{Kris~M. White}, \bibinfo{person}{Matthew~J. O’Meara},
  \bibinfo{person}{Veronica~V. Rezelj}, \bibinfo{person}{Jeffrey~Z. Guo},
  \bibinfo{person}{Danielle~L. Swaney}, \bibinfo{person}{Tia~A. Tummino},
  \bibinfo{person}{Ruth Huettenhain}, \bibinfo{person}{Robyn~M. Kaake},
  \bibinfo{person}{Alicia~L. Richards}, \bibinfo{person}{Beril Tutuncuoglu},
  \bibinfo{person}{Helene Foussard}, \bibinfo{person}{Jyoti Batra},
  \bibinfo{person}{Kelsey Haas}, \bibinfo{person}{Maya Modak},
  \bibinfo{person}{Minkyu Kim}, \bibinfo{person}{Paige Haas},
  \bibinfo{person}{Benjamin~J. Polacco}, \bibinfo{person}{Hannes Braberg},
  \bibinfo{person}{Jacqueline~M. Fabius}, \bibinfo{person}{Manon Eckhardt},
  \bibinfo{person}{Margaret Soucheray}, \bibinfo{person}{Melanie~J. Bennett},
  \bibinfo{person}{Merve Cakir}, \bibinfo{person}{Michael~J. McGregor},
  \bibinfo{person}{Qiongyu Li}, \bibinfo{person}{Bjoern Meyer},
  \bibinfo{person}{Ferdinand Roesch}, \bibinfo{person}{Thomas Vallet},
  \bibinfo{person}{Alice Mac~Kain}, \bibinfo{person}{Lisa Miorin},
  \bibinfo{person}{Elena Moreno}, \bibinfo{person}{Zun Zar~Chi Naing},
  \bibinfo{person}{Yuan Zhou}, \bibinfo{person}{Shiming Peng},
  \bibinfo{person}{Ying Shi}, \bibinfo{person}{Ziyang Zhang},
  \bibinfo{person}{Wenqi Shen}, \bibinfo{person}{Ilsa~T. Kirby},
  \bibinfo{person}{James~E. Melnyk}, \bibinfo{person}{John~S. Chorba},
  \bibinfo{person}{Kevin Lou}, \bibinfo{person}{Shizhong~A. Dai},
  \bibinfo{person}{Inigo Barrio-Hernandez}, \bibinfo{person}{Danish Memon},
  \bibinfo{person}{Claudia Hernandez-Armenta}, \bibinfo{person}{Jiankun Lyu},
  \bibinfo{person}{Christopher J.~P. Mathy}, \bibinfo{person}{Tina Perica},
  \bibinfo{person}{Kala~B. Pilla}, \bibinfo{person}{Sai~J. Ganesan},
  \bibinfo{person}{Daniel~J. Saltzberg}, \bibinfo{person}{Ramachandran Rakesh},
  \bibinfo{person}{Xi Liu}, \bibinfo{person}{Sara~B. Rosenthal},
  \bibinfo{person}{Lorenzo Calviello}, \bibinfo{person}{Srivats
  Venkataramanan}, \bibinfo{person}{Jose Liboy-Lugo}, \bibinfo{person}{Yizhu
  Lin}, \bibinfo{person}{Xi-Ping Huang}, \bibinfo{person}{YongFeng Liu},
  \bibinfo{person}{Stephanie~A. Wankowicz}, \bibinfo{person}{Markus Bohn},
  \bibinfo{person}{Maliheh Safari}, \bibinfo{person}{Fatima~S. Ugur},
  \bibinfo{person}{Cassandra Koh}, \bibinfo{person}{Nastaran~Sadat Savar},
  \bibinfo{person}{Quang~Dinh Tran}, \bibinfo{person}{Djoshkun Shengjuler},
  \bibinfo{person}{Sabrina~J. Fletcher}, \bibinfo{person}{Michael~C. O’Neal},
  \bibinfo{person}{Yiming Cai}, \bibinfo{person}{Jason C.~J. Chang},
  \bibinfo{person}{David~J. Broadhurst}, \bibinfo{person}{Saker Klippsten},
  \bibinfo{person}{Phillip~P. Sharp}, \bibinfo{person}{Nicole~A. Wenzell},
  \bibinfo{person}{Duygu Kuzuoglu}, \bibinfo{person}{Hao-Yuan Wang},
  \bibinfo{person}{Raphael Trenker}, \bibinfo{person}{Janet~M. Young},
  \bibinfo{person}{Devin~A. Cavero}, \bibinfo{person}{Joseph Hiatt},
  \bibinfo{person}{Theodore~L. Roth}, \bibinfo{person}{Ujjwal Rathore},
  \bibinfo{person}{Advait Subramanian}, \bibinfo{person}{Julia Noack},
  \bibinfo{person}{Mathieu Hubert}, \bibinfo{person}{Robert~M. Stroud},
  \bibinfo{person}{Alan~D. Frankel}, \bibinfo{person}{Oren~S. Rosenberg},
  \bibinfo{person}{Kliment~A. Verba}, \bibinfo{person}{David~A. Agard},
  \bibinfo{person}{Melanie Ott}, \bibinfo{person}{Michael Emerman},
  \bibinfo{person}{Natalia Jura}, {et~al\mbox{.}}}
  \bibinfo{year}{2020}\natexlab{}.
\newblock \showarticletitle{A SARS-CoV-2 protein interaction map reveals
  targets for drug repurposing}.
\newblock \bibinfo{journal}{\emph{Nature}} (\bibinfo{year}{2020}).
\newblock
\showISSN{1476-4687}
\urldef\tempurl%
\url{https://doi.org/10.1038/s41586-020-2286-9}
\showDOI{\tempurl}


\bibitem[\protect\citeauthoryear{Gyebi, Ogunro, Adegunloye, Ogunyemi, and
  Afolabi}{Gyebi et~al\mbox{.}}{2020}]%
        {3CLforCOVID}
\bibfield{author}{\bibinfo{person}{Gideon~A. Gyebi},
  \bibinfo{person}{Olalekan~B. Ogunro}, \bibinfo{person}{Adegbenro~P.
  Adegunloye}, \bibinfo{person}{Oludare~M. Ogunyemi}, {and}
  \bibinfo{person}{Saheed~O. Afolabi}.} \bibinfo{year}{2020}\natexlab{}.
\newblock \showarticletitle{Potential inhibitors of coronavirus
  3-chymotrypsin-like protease (3CL(pro)): an in silico screening of alkaloids
  and terpenoids from African medicinal plants}.
\newblock \bibinfo{journal}{\emph{Journal of Biomolecular Structure and
  Dynamics}} (\bibinfo{year}{2020}), \bibinfo{pages}{1--13}.
\newblock
\showISSN{1538-0254 0739-1102}
\urldef\tempurl%
\url{https://doi.org/10.1080/07391102.2020.1764868}
\showDOI{\tempurl}


\bibitem[\protect\citeauthoryear{Harrison}{Harrison}{2020}]%
        {harrison2020coronavirus}
\bibfield{author}{\bibinfo{person}{C Harrison}.}
  \bibinfo{year}{2020}\natexlab{}.
\newblock \showarticletitle{Coronavirus puts drug repurposing on the fast
  track.}
\newblock \bibinfo{journal}{\emph{Nature Biotechnology}} \bibinfo{volume}{38},
  \bibinfo{number}{4} (\bibinfo{year}{2020}), \bibinfo{pages}{379}.
\newblock


\bibitem[\protect\citeauthoryear{He, Heidemeyer, Ban, Cherkasov, and Ester}{He
  et~al\mbox{.}}{2017}]%
        {he2017simboost}
\bibfield{author}{\bibinfo{person}{Tong He}, \bibinfo{person}{Marten
  Heidemeyer}, \bibinfo{person}{Fuqiang Ban}, \bibinfo{person}{Artem
  Cherkasov}, {and} \bibinfo{person}{Martin Ester}.}
  \bibinfo{year}{2017}\natexlab{}.
\newblock \showarticletitle{SimBoost: a read-across approach for predicting
  drug--target binding affinities using gradient boosting machines}.
\newblock \bibinfo{journal}{\emph{Journal of Cheminformatics}}
  \bibinfo{volume}{9}, \bibinfo{number}{1} (\bibinfo{year}{2017}),
  \bibinfo{pages}{1--14}.
\newblock


\bibitem[\protect\citeauthoryear{Horwood and Noutahi}{Horwood and
  Noutahi}{2020}]%
        {horwood2020molecular}
\bibfield{author}{\bibinfo{person}{Julien Horwood} {and}
  \bibinfo{person}{Emmanuel Noutahi}.} \bibinfo{year}{2020}\natexlab{}.
\newblock \showarticletitle{Molecular Design in Synthetically Accessible
  Chemical Space via Deep Reinforcement Learning}.
\newblock \bibinfo{journal}{\emph{arXiv preprint arXiv:2004.14308}}
  (\bibinfo{year}{2020}).
\newblock


\bibitem[\protect\citeauthoryear{Huang, Fu, Xiao, Glass, and Sun}{Huang
  et~al\mbox{.}}{2020}]%
        {huang2020deeppurpose}
\bibfield{author}{\bibinfo{person}{Kexin Huang}, \bibinfo{person}{Tianfan Fu},
  \bibinfo{person}{Cao Xiao}, \bibinfo{person}{Lucas Glass}, {and}
  \bibinfo{person}{Jimeng Sun}.} \bibinfo{year}{2020}\natexlab{}.
\newblock \showarticletitle{DeepPurpose: a Deep Learning Based Drug Repurposing
  Toolkit}.
\newblock \bibinfo{journal}{\emph{arXiv preprint arXiv:2004.08919}}
  (\bibinfo{year}{2020}).
\newblock


\bibitem[\protect\citeauthoryear{{Jin}, {Barzilay}, and {Jaakkola}}{{Jin}
  et~al\mbox{.}}{2020}]%
        {JTVAE-multi}
\bibfield{author}{\bibinfo{person}{Wengong {Jin}}, \bibinfo{person}{Regina
  {Barzilay}}, {and} \bibinfo{person}{Tommi {Jaakkola}}.}
  \bibinfo{year}{2020}\natexlab{}.
\newblock \showarticletitle{{Multi-Objective Molecule Generation using
  Interpretable Substructures}}.
\newblock \bibinfo{journal}{\emph{arXiv e-prints}}, Article
  \bibinfo{articleno}{arXiv:2002.03244} (\bibinfo{date}{Feb.}
  \bibinfo{year}{2020}), \bibinfo{numpages}{arXiv:2002.03244}~pages.
\newblock
\showeprint[arxiv]{cs.LG/2002.03244}


\bibitem[\protect\citeauthoryear{Jin, Barzilay, and Jaakkola}{Jin
  et~al\mbox{.}}{2018}]%
        {JTVAE}
\bibfield{author}{\bibinfo{person}{Wengong Jin}, \bibinfo{person}{Regina
  Barzilay}, {and} \bibinfo{person}{Tommi~S. Jaakkola}.}
  \bibinfo{year}{2018}\natexlab{}.
\newblock \showarticletitle{Junction Tree Variational Autoencoder for Molecular
  Graph Generation}.
\newblock \bibinfo{journal}{\emph{CoRR}}  \bibinfo{volume}{abs/1802.04364}
  (\bibinfo{year}{2018}).
\newblock
\showeprint[arxiv]{1802.04364}
\urldef\tempurl%
\url{http://arxiv.org/abs/1802.04364}
\showURL{%
\tempurl}


\bibitem[\protect\citeauthoryear{John, Phillips, Kemper, Wilson, Guan, Crowley,
  Nimlos, and Larsen}{John et~al\mbox{.}}{2019}]%
        {stjohn2019mpnnpolymer}
\bibfield{author}{\bibinfo{person}{Peter C.~St. John}, \bibinfo{person}{Caleb
  Phillips}, \bibinfo{person}{Travis~W. Kemper}, \bibinfo{person}{A.~Nolan
  Wilson}, \bibinfo{person}{Yanfei Guan}, \bibinfo{person}{Michael~F. Crowley},
  \bibinfo{person}{Mark~R. Nimlos}, {and} \bibinfo{person}{Ross~E. Larsen}.}
  \bibinfo{year}{2019}\natexlab{}.
\newblock \showarticletitle{Message-passing neural networks for high-throughput
  polymer screening}.
\newblock \bibinfo{journal}{\emph{Journal of Chemical Physics}}
  \bibinfo{volume}{150}, \bibinfo{number}{23} (\bibinfo{date}{June}
  \bibinfo{year}{2019}), \bibinfo{pages}{234111}.
\newblock
\urldef\tempurl%
\url{https://doi.org/10.1063/1.5099132}
\showDOI{\tempurl}


\bibitem[\protect\citeauthoryear{Khemchandani, O'Hagan, Samanta, Swainston,
  Roberts, Bollegala, and Kell}{Khemchandani et~al\mbox{.}}{2020}]%
        {khemchandani2020deepgraphmol}
\bibfield{author}{\bibinfo{person}{Yash Khemchandani}, \bibinfo{person}{Steve
  O'Hagan}, \bibinfo{person}{Soumitra Samanta}, \bibinfo{person}{Neil
  Swainston}, \bibinfo{person}{Timothy~J Roberts}, \bibinfo{person}{Danushka
  Bollegala}, {and} \bibinfo{person}{Douglas~B Kell}.}
  \bibinfo{year}{2020}\natexlab{}.
\newblock \showarticletitle{DeepGraphMol, a multi-objective, computational
  strategy for generating molecules with desirable properties: a graph
  convolution and reinforcement learning approach}.
\newblock \bibinfo{journal}{\emph{bioRxiv}} (\bibinfo{year}{2020}).
\newblock


\bibitem[\protect\citeauthoryear{Kumar and Zhang}{Kumar and Zhang}{2018}]%
        {Similarity2018}
\bibfield{author}{\bibinfo{person}{Ashutosh Kumar} {and} \bibinfo{person}{Kam
  Y.~J. Zhang}.} \bibinfo{year}{2018}\natexlab{}.
\newblock \showarticletitle{Advances in the Development of Shape Similarity
  Methods and Their Application in Drug Discovery}.
\newblock \bibinfo{journal}{\emph{Frontiers in chemistry}}  \bibinfo{volume}{6}
  (\bibinfo{year}{2018}), \bibinfo{pages}{315--315}.
\newblock
\showISSN{2296-2646}
\urldef\tempurl%
\url{https://doi.org/10.3389/fchem.2018.00315}
\showDOI{\tempurl}


\bibitem[\protect\citeauthoryear{Kusner, Paige, and Hernández-Lobato}{Kusner
  et~al\mbox{.}}{2017}]%
        {GrammarBO}
\bibfield{author}{\bibinfo{person}{Matt~J. Kusner}, \bibinfo{person}{Brooks
  Paige}, {and} \bibinfo{person}{José~Miguel Hernández-Lobato}.}
  \bibinfo{year}{2017}\natexlab{}.
\newblock \bibinfo{title}{Grammar Variational Autoencoder}.
\newblock
\newblock
\showeprint[arxiv]{stat.ML/1703.01925}


\bibitem[\protect\citeauthoryear{Lauritzen and Spiegelhalter}{Lauritzen and
  Spiegelhalter}{1988}]%
        {lauritzen1988local}
\bibfield{author}{\bibinfo{person}{Steffen~L Lauritzen} {and}
  \bibinfo{person}{David~J Spiegelhalter}.} \bibinfo{year}{1988}\natexlab{}.
\newblock \showarticletitle{Local computations with probabilities on graphical
  structures and their application to expert systems}.
\newblock \bibinfo{journal}{\emph{Journal of the Royal Statistical Society:
  Series B (Methodological)}} \bibinfo{volume}{50}, \bibinfo{number}{2}
  (\bibinfo{year}{1988}), \bibinfo{pages}{157--194}.
\newblock


\bibitem[\protect\citeauthoryear{Li, Zhang, and Liu}{Li et~al\mbox{.}}{2018}]%
        {Li2018GraphGen}
\bibfield{author}{\bibinfo{person}{Yibo Li}, \bibinfo{person}{Liangren Zhang},
  {and} \bibinfo{person}{Zhenming Liu}.} \bibinfo{year}{2018}\natexlab{}.
\newblock \showarticletitle{Multi-objective de novo drug design with
  conditional graph generative model}.
\newblock \bibinfo{journal}{\emph{Journal of Cheminformatics}}
  \bibinfo{volume}{10}, \bibinfo{number}{1} (\bibinfo{year}{2018}),
  \bibinfo{pages}{33}.
\newblock
\showISSN{1758-2946}
\urldef\tempurl%
\url{https://doi.org/10.1186/s13321-018-0287-6}
\showDOI{\tempurl}


\bibitem[\protect\citeauthoryear{Lipinski}{Lipinski}{2004}]%
        {lipinski2004lead}
\bibfield{author}{\bibinfo{person}{Christopher~A Lipinski}.}
  \bibinfo{year}{2004}\natexlab{}.
\newblock \showarticletitle{Lead-and drug-like compounds: the rule-of-five
  revolution}.
\newblock \bibinfo{journal}{\emph{Drug Discovery Today: Technologies}}
  \bibinfo{volume}{1}, \bibinfo{number}{4} (\bibinfo{year}{2004}),
  \bibinfo{pages}{337--341}.
\newblock


\bibitem[\protect\citeauthoryear{Lu}{Lu}{2020}]%
        {COVID-options}
\bibfield{author}{\bibinfo{person}{Hongzhou Lu}.}
  \bibinfo{year}{2020}\natexlab{}.
\newblock \showarticletitle{Drug treatment options for the 2019-new coronavirus
  (2019-nCoV)}.
\newblock \bibinfo{journal}{\emph{BioScience Trends}} \bibinfo{volume}{14},
  \bibinfo{number}{1} (\bibinfo{year}{2020}), \bibinfo{pages}{69--71}.
\newblock
\urldef\tempurl%
\url{https://doi.org/10.5582/bst.2020.01020}
\showDOI{\tempurl}


\bibitem[\protect\citeauthoryear{Maggiora, Vogt, Stumpfe, and
  Bajorath}{Maggiora et~al\mbox{.}}{2014}]%
        {Similarity2014}
\bibfield{author}{\bibinfo{person}{Gerald Maggiora}, \bibinfo{person}{Martin
  Vogt}, \bibinfo{person}{Dagmar Stumpfe}, {and} \bibinfo{person}{Jurgen
  Bajorath}.} \bibinfo{year}{2014}\natexlab{}.
\newblock \showarticletitle{Molecular similarity in medicinal chemistry:
  miniperspective}.
\newblock \bibinfo{journal}{\emph{Journal of medicinal chemistry}}
  \bibinfo{volume}{57}, \bibinfo{number}{8} (\bibinfo{year}{2014}),
  \bibinfo{pages}{3186--3204}.
\newblock


\bibitem[\protect\citeauthoryear{Mnih, Kavukcuoglu, Silver, Graves, Antonoglou,
  Wierstra, and Riedmiller}{Mnih et~al\mbox{.}}{2013}]%
        {DQN}
\bibfield{author}{\bibinfo{person}{Volodymyr Mnih}, \bibinfo{person}{Koray
  Kavukcuoglu}, \bibinfo{person}{David Silver}, \bibinfo{person}{Alex Graves},
  \bibinfo{person}{Ioannis Antonoglou}, \bibinfo{person}{Daan Wierstra}, {and}
  \bibinfo{person}{Martin Riedmiller}.} \bibinfo{year}{2013}\natexlab{}.
\newblock \showarticletitle{Playing atari with deep reinforcement learning}.
\newblock \bibinfo{journal}{\emph{arXiv preprint arXiv:1312.5602}}
  (\bibinfo{year}{2013}).
\newblock


\bibitem[\protect\citeauthoryear{Patankar and Jurs}{Patankar and Jurs}{2000}]%
        {IC50-QSAR}
\bibfield{author}{\bibinfo{person}{S.~J. Patankar} {and} \bibinfo{person}{P.~C.
  Jurs}.} \bibinfo{year}{2000}\natexlab{}.
\newblock \showarticletitle{Prediction of IC50 Values for ACAT Inhibitors from
  Molecular Structure}.
\newblock \bibinfo{journal}{\emph{Journal of Chemical Information and Computer
  Sciences}} \bibinfo{volume}{40}, \bibinfo{number}{3} (\bibinfo{year}{2000}),
  \bibinfo{pages}{706--723}.
\newblock
\showISSN{0095-2338}
\urldef\tempurl%
\url{https://doi.org/10.1021/ci990125r}
\showDOI{\tempurl}


\bibitem[\protect\citeauthoryear{Rogers and Hahn}{Rogers and Hahn}{2010}]%
        {rogers2010extended}
\bibfield{author}{\bibinfo{person}{David Rogers} {and} \bibinfo{person}{Mathew
  Hahn}.} \bibinfo{year}{2010}\natexlab{}.
\newblock \showarticletitle{Extended-connectivity fingerprints}.
\newblock \bibinfo{journal}{\emph{Journal of Chemical Information and
  Modeling}} \bibinfo{volume}{50}, \bibinfo{number}{5} (\bibinfo{year}{2010}),
  \bibinfo{pages}{742--754}.
\newblock


\bibitem[\protect\citeauthoryear{Samanta, Abir, Jana, Chattaraj, Ganguly, and
  Rodriguez}{Samanta et~al\mbox{.}}{2019}]%
        {samanta2019nevae}
\bibfield{author}{\bibinfo{person}{Bidisha Samanta}, \bibinfo{person}{DE Abir},
  \bibinfo{person}{Gourhari Jana}, \bibinfo{person}{Pratim~Kumar Chattaraj},
  \bibinfo{person}{Niloy Ganguly}, {and} \bibinfo{person}{Manuel~Gomez
  Rodriguez}.} \bibinfo{year}{2019}\natexlab{}.
\newblock \showarticletitle{Nevae: A deep generative model for molecular
  graphs}. In \bibinfo{booktitle}{\emph{AAAI Conference on Artificial
  Intelligence}}, Vol.~\bibinfo{volume}{33}. \bibinfo{pages}{1110--1117}.
\newblock


\bibitem[\protect\citeauthoryear{Scarselli, Gori, Tsoi, Hagenbuchner, and
  Monfardini}{Scarselli et~al\mbox{.}}{2008}]%
        {scarselli2008graph}
\bibfield{author}{\bibinfo{person}{Franco Scarselli}, \bibinfo{person}{Marco
  Gori}, \bibinfo{person}{Ah~Chung Tsoi}, \bibinfo{person}{Markus
  Hagenbuchner}, {and} \bibinfo{person}{Gabriele Monfardini}.}
  \bibinfo{year}{2008}\natexlab{}.
\newblock \showarticletitle{The graph neural network model}.
\newblock \bibinfo{journal}{\emph{IEEE Transactions on Neural Networks}}
  \bibinfo{volume}{20}, \bibinfo{number}{1} (\bibinfo{year}{2008}),
  \bibinfo{pages}{61--80}.
\newblock


\bibitem[\protect\citeauthoryear{Sebaugh}{Sebaugh}{2011}]%
        {sebaugh2011guidelines}
\bibfield{author}{\bibinfo{person}{JL Sebaugh}.}
  \bibinfo{year}{2011}\natexlab{}.
\newblock \showarticletitle{Guidelines for accurate EC50/IC50 estimation}.
\newblock \bibinfo{journal}{\emph{Pharmaceutical statistics}}
  \bibinfo{volume}{10}, \bibinfo{number}{2} (\bibinfo{year}{2011}),
  \bibinfo{pages}{128--134}.
\newblock


\bibitem[\protect\citeauthoryear{Sivaraman, Jackson, Sanchez-Lengeling,
  Vasquez-Mayagoitia, Aspuru-Guzik, Vishwanath, and de~Pablo}{Sivaraman
  et~al\mbox{.}}{2020}]%
        {selfiegeneration2020}
\bibfield{author}{\bibinfo{person}{Ganesh Sivaraman}, \bibinfo{person}{Nicholas
  Jackson}, \bibinfo{person}{Benjamin Sanchez-Lengeling},
  \bibinfo{person}{Alvaro Vasquez-Mayagoitia}, \bibinfo{person}{Alan
  Aspuru-Guzik}, \bibinfo{person}{Venkatram Vishwanath}, {and}
  \bibinfo{person}{Juan de Pablo}.} \bibinfo{year}{2020}\natexlab{}.
\newblock \showarticletitle{A machine learning workflow for molecular analysis:
  application to melting points}.
\newblock \bibinfo{journal}{\emph{Machine Learning: Science and Technology}}
  (\bibinfo{year}{2020}).
\newblock


\bibitem[\protect\citeauthoryear{Smith and Smith}{Smith and Smith}{2020}]%
        {SummitDocking}
\bibfield{author}{\bibinfo{person}{Micholas Smith} {and}
  \bibinfo{person}{Jeremy~C. Smith}.} \bibinfo{year}{2020}\natexlab{}.
\newblock \bibinfo{title}{Repurposing Therapeutics for COVID-19:
  Supercomputer-Based Docking to the SARS-CoV-2 Viral Spike Protein and Viral
  Spike Protein-Human ACE2 Interface}.
\newblock
\newblock
\urldef\tempurl%
\url{https://doi.org/10.26434/chemrxiv.11871402.v3}
\showDOI{\tempurl}


\bibitem[\protect\citeauthoryear{St{\aa}hl, Falkman, Karlsson, Mathiason, and
  Bostrom}{St{\aa}hl et~al\mbox{.}}{2019}]%
        {staahl2019deep}
\bibfield{author}{\bibinfo{person}{Niclas St{\aa}hl},
  \bibinfo{person}{G\"{o}ran Falkman}, \bibinfo{person}{Alexander Karlsson},
  \bibinfo{person}{Gunnar Mathiason}, {and} \bibinfo{person}{Jonas Bostrom}.}
  \bibinfo{year}{2019}\natexlab{}.
\newblock \showarticletitle{Deep reinforcement learning for multiparameter
  optimization in de novo drug design}.
\newblock \bibinfo{journal}{\emph{Journal of Chemical Information and
  Modeling}} \bibinfo{volume}{59}, \bibinfo{number}{7} (\bibinfo{year}{2019}),
  \bibinfo{pages}{3166--3176}.
\newblock


\bibitem[\protect\citeauthoryear{Thafar, Raies, Albaradei, Essack, and
  Bajic}{Thafar et~al\mbox{.}}{2019}]%
        {thafar2019comparison}
\bibfield{author}{\bibinfo{person}{Maha Thafar}, \bibinfo{person}{Arwa~Bin
  Raies}, \bibinfo{person}{Somayah Albaradei}, \bibinfo{person}{Magbubah
  Essack}, {and} \bibinfo{person}{Vladimir~B Bajic}.}
  \bibinfo{year}{2019}\natexlab{}.
\newblock \showarticletitle{Comparison Study of Computational Prediction Tools
  for Drug-Target Binding Affinities}.
\newblock \bibinfo{journal}{\emph{Frontiers in Chemistry}}  \bibinfo{volume}{7}
  (\bibinfo{year}{2019}).
\newblock


\bibitem[\protect\citeauthoryear{Waring}{Waring}{2010}]%
        {logP2010}
\bibfield{author}{\bibinfo{person}{Michael~J Waring}.}
  \bibinfo{year}{2010}\natexlab{}.
\newblock \showarticletitle{Lipophilicity in drug discovery}.
\newblock \bibinfo{journal}{\emph{Expert Opinion on Drug Discovery}}
  \bibinfo{volume}{5}, \bibinfo{number}{3} (\bibinfo{year}{2010}),
  \bibinfo{pages}{235--248}.
\newblock
\urldef\tempurl%
\url{https://doi.org/10.1517/17460441003605098}
\showDOI{\tempurl}
\showeprint{https://doi.org/10.1517/17460441003605098}
\newblock
\shownote{PMID: 22823020.}


\bibitem[\protect\citeauthoryear{Weininger}{Weininger}{1990}]%
        {weininger1990smiles}
\bibfield{author}{\bibinfo{person}{David Weininger}.}
  \bibinfo{year}{1990}\natexlab{}.
\newblock \showarticletitle{SMILES. 3. DEPICT. Graphical depiction of chemical
  structures}.
\newblock \bibinfo{journal}{\emph{Journal of Chemical Information and Computer
  Sciences}} \bibinfo{volume}{30}, \bibinfo{number}{3} (\bibinfo{year}{1990}),
  \bibinfo{pages}{237--243}.
\newblock


\bibitem[\protect\citeauthoryear{You, Liu, Ying, Pande, and Leskovec}{You
  et~al\mbox{.}}{2018}]%
        {GCPN}
\bibfield{author}{\bibinfo{person}{Jiaxuan You}, \bibinfo{person}{Bowen Liu},
  \bibinfo{person}{Zhitao Ying}, \bibinfo{person}{Vijay Pande}, {and}
  \bibinfo{person}{Jure Leskovec}.} \bibinfo{year}{2018}\natexlab{}.
\newblock \showarticletitle{Graph convolutional policy network for
  goal-directed molecular graph generation}. In
  \bibinfo{booktitle}{\emph{Advances in Neural Information Processing
  Systems}}. \bibinfo{pages}{6410--6421}.
\newblock


\bibitem[\protect\citeauthoryear{Zhang and Wilkinson}{Zhang and
  Wilkinson}{2007}]%
        {Beyond5}
\bibfield{author}{\bibinfo{person}{Ming-Qiang Zhang} {and}
  \bibinfo{person}{Barrie Wilkinson}.} \bibinfo{year}{2007}\natexlab{}.
\newblock \showarticletitle{Drug discovery beyond the ‘rule-of-five’}.
\newblock \bibinfo{journal}{\emph{Current Opinion in Biotechnology}}
  \bibinfo{volume}{18}, \bibinfo{number}{6} (\bibinfo{year}{2007}),
  \bibinfo{pages}{478 -- 488}.
\newblock
\showISSN{0958-1669}
\urldef\tempurl%
\url{https://doi.org/10.1016/j.copbio.2007.10.005}
\showDOI{\tempurl}
\newblock
\shownote{Chemical biotechnology / Pharmaceutical biotechnology.}


\bibitem[\protect\citeauthoryear{Zhou, Hou, Shen, Huang, Martin, and
  Cheng}{Zhou et~al\mbox{.}}{2020}]%
        {COVID-repurpose-nature}
\bibfield{author}{\bibinfo{person}{Yadi Zhou}, \bibinfo{person}{Yuan Hou},
  \bibinfo{person}{Jiayu Shen}, \bibinfo{person}{Yin Huang},
  \bibinfo{person}{William Martin}, {and} \bibinfo{person}{Feixiong Cheng}.}
  \bibinfo{year}{2020}\natexlab{}.
\newblock \showarticletitle{Network-based drug repurposing for novel
  coronavirus 2019-nCoV/SARS-CoV-2}.
\newblock \bibinfo{journal}{\emph{Cell Discovery}} \bibinfo{volume}{6},
  \bibinfo{number}{1} (\bibinfo{year}{2020}), \bibinfo{pages}{14}.
\newblock
\showISSN{2056-5968}
\urldef\tempurl%
\url{https://doi.org/10.1038/s41421-020-0153-3}
\showDOI{\tempurl}


\bibitem[\protect\citeauthoryear{Zhou, Kearnes, Li, Zare, and Riley}{Zhou
  et~al\mbox{.}}{2019}]%
        {MolDQN}
\bibfield{author}{\bibinfo{person}{Zhenpeng Zhou}, \bibinfo{person}{Steven
  Kearnes}, \bibinfo{person}{Li Li}, \bibinfo{person}{Richard~N. Zare}, {and}
  \bibinfo{person}{Patrick Riley}.} \bibinfo{year}{2019}\natexlab{}.
\newblock \showarticletitle{Optimization of Molecules via Deep Reinforcement
  Learning}.
\newblock \bibinfo{journal}{\emph{Scientific Reports}} \bibinfo{volume}{9},
  \bibinfo{number}{1} (\bibinfo{year}{2019}), \bibinfo{pages}{10752}.
\newblock
\showISSN{2045-2322}
\urldef\tempurl%
\url{https://doi.org/10.1038/s41598-019-47148-x}
\showDOI{\tempurl}


\end{thebibliography}

% \clearpage
\renewcommand{\thefigure}{S\arabic{figure}}
\setcounter{figure}{0} 
\renewcommand{\thetable}{S\arabic{table}}
\setcounter{table}{0} 
\setcounter{page}{1}
\renewcommand{\thepage}{S\arabic{page}}

\clearpage
\onecolumn
\section*{Supplementary Material}

\subsection{Dataset Description}
\label{sec:Dataset Preparation}

\textbf{Preparation.} We prepared and assembled the protease datasets with molecules active against various protease in enzymatic assays filtered from experimentally pharmacology data such as CheMBL, BindingDB, and ToxCat \cite{bento2014chembl}. The database was filtered out with the IC\textsubscript{50} activity standard types and their potency. Molecules with size larger than 1000 Dalton were removed due to the limitation of the representation of large molecules in cheminformatics. We also filtered out non-drug like molecules containing metals and polypeptides.  The curated data was standardized using the logarithmic scale –log10 of a numeric value in nM for all compounds. We used the mean average for a molecule with more than one IC\textsubscript{50} value. The resulting dataset contains 6545 unique molecules accompanied by their SMILES strings and experimental IC\textsubscript{50} values.   

\begin{figure*}[htbp!] \centering
  \includegraphics[clip, trim=0.5cm 6.5cm 0.15cm 6.5cm, width=0.75\textwidth]{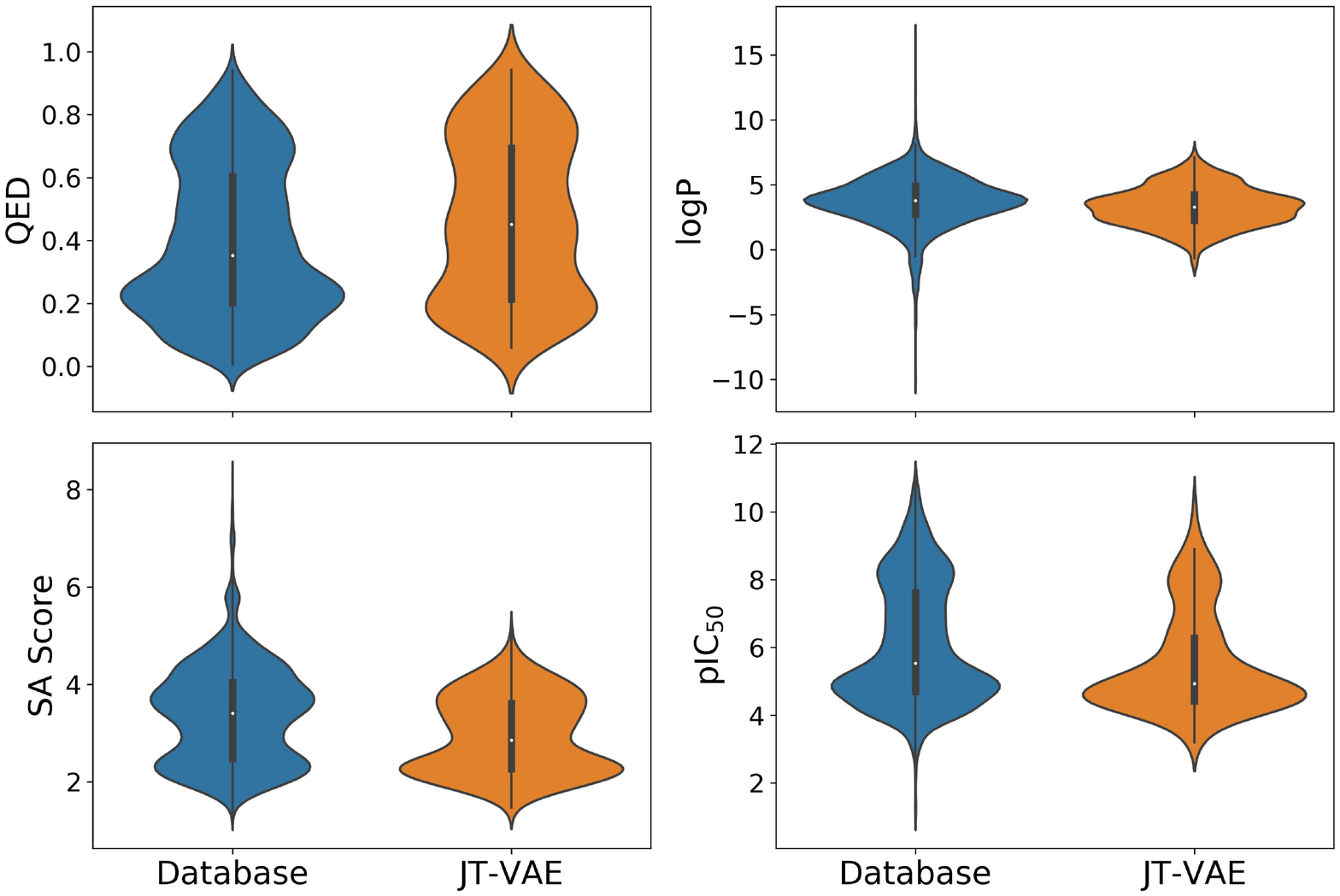}
  \caption{We generated 1,000 molecules using the trained JT-VAE, of which 560 were unique. The figure shows the comparison of the QED, logP, SA score, and pIC\textsubscript{50} of compounds in the database and those generated by the JT-VAE. The JT-VAE reproduced the range of values present in the database, minus outliers. The similar values indicate that the JT-VAE is able to reproduce the wide range of structures present in the database. The pIC\textsubscript{50} values for generated molecules were estimated by our MPNN.} \label{fig:DBvVAE}
\end{figure*}

\textbf{Quantitative Characterization.} We computed the metrics included in our scoring functions for each molecule in the database. The range of values can be seen in Figure \ref{fig:DBvVAE}. The highest pIC\textsubscript{50} is 10.89, while the lowest pIC\textsubscript{50} is 1.22. The most common pIC\textsubscript{50} is 4.0, which is shared by 320 structures, and the vast majority of structures (91.5\%) have pIC\textsubscript{50} values greater than 4.0. LogP values range from -10.36 to 16.65 in a near Gaussian distribution with a mean of 3.70; 77.0\% of all structures meet the requirement of Lipinski's Rule of 5 that logP be no greater than 5. QED values range from 0.01 to 0.94, while SA values range from 1.35 to 8.24, with 96.3\% being below 5.

\begin{figure*}[htbp!] \centering
  \includegraphics[clip, trim=0.15cm 5.2cm 1.5cm 5.2cm, width=0.48\textwidth]{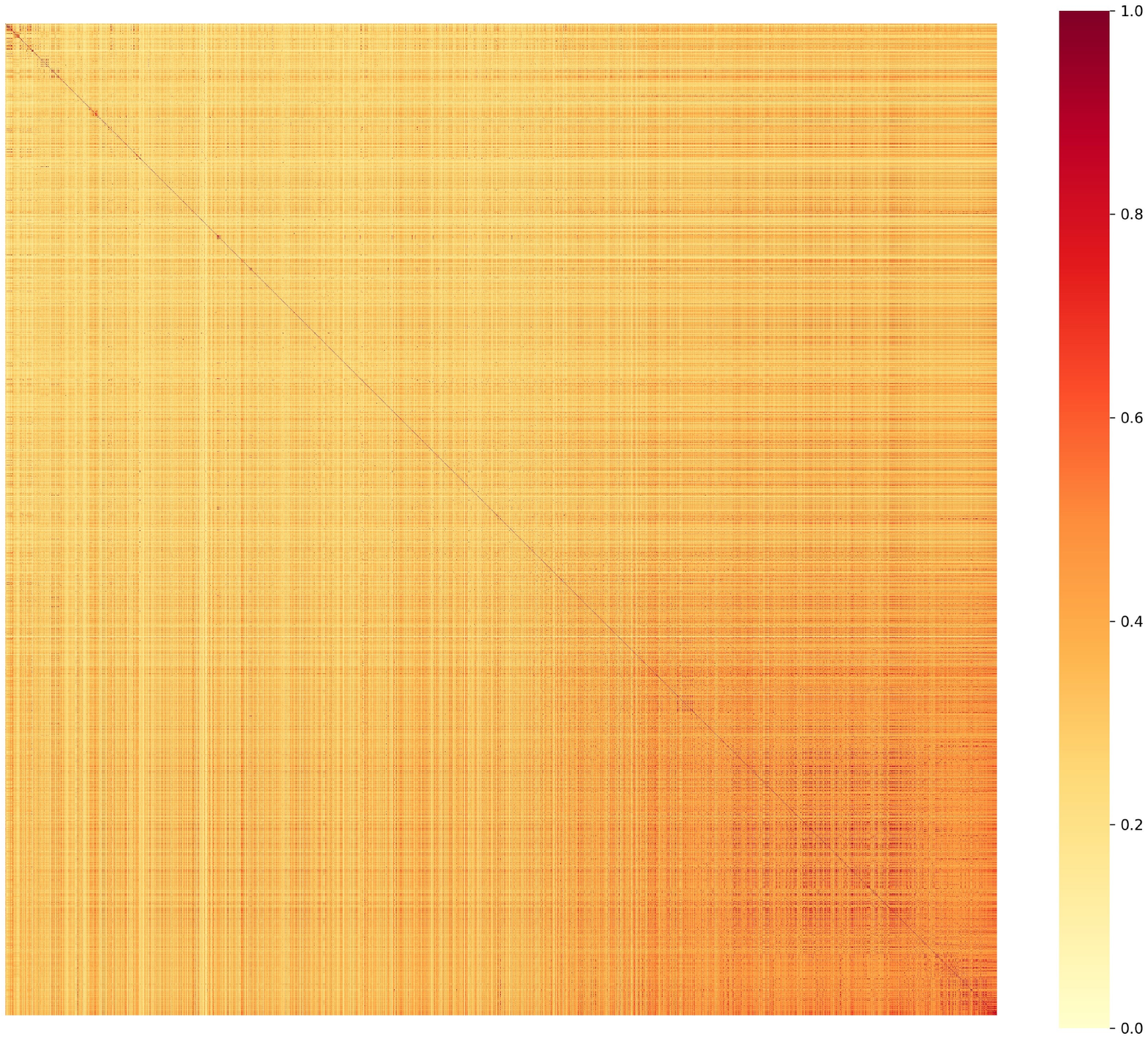}
  \caption{Heat map of the Tanimoto similarity between all compounds in the database. Entries are ordered in increasing pIC\textsubscript{50}. The Tanimoto similarity is generally higher among structures with high pIC\textsubscript{50} (located towards bottom right of the matrix), indicating the importance of considering structural similarity in drug discovery.} \label{fig:database_similarity}
\end{figure*}

We computed the Tanimoto similarity for all pairs of compounds to gain insight into the structural diversity of molecules in our database (Figure \ref{fig:database_similarity}). The entries in the matrix were ordered in increasing pIC\textsubscript{50} values.  The similarity is represented by the color bar, with yellow representing low similarity (0) and red high similarity (1). We observe that structures tend to become more similar to their neighbors as pIC\textsubscript{50} increases, indicating that compounds with high pIC\textsubscript{50} values tend to be structurally similar, supporting the consideration of molecular similarity during drug discovery.

\clearpage
\begin{table*}[]
    \centering
    \caption{$R^2$ scores for MPNN models trained to predict the pIC\textsubscript{50} of drugs
    from their molecular structure. Each model was trained using a different readout
    function (columns) to combine atomic contributions to pIC\textsubscript{50} or 
    to create a single molecular fingerprint. Bold indicates the model used in our
    experiments and underscore indicates the best-performing model.}
    \begin{tabular}{lrr}
    \toprule
     &  atomic &  molecular \\
    readout   &         &            \\
    \midrule
    attention &    \textbf{0.57} &       0.56 \\
    max       &    0.61 &       \underline{0.71} \\
    mean      &    0.57 &       0.61 \\
    softmax   &    0.61 &       0.70 \\
    sum       &    0.51 &       0.54 \\
    \bottomrule
    \end{tabular}
    \label{tab:mpnn_results}
\end{table*}

\begin{figure*}[htbp!] \centering
  \includegraphics{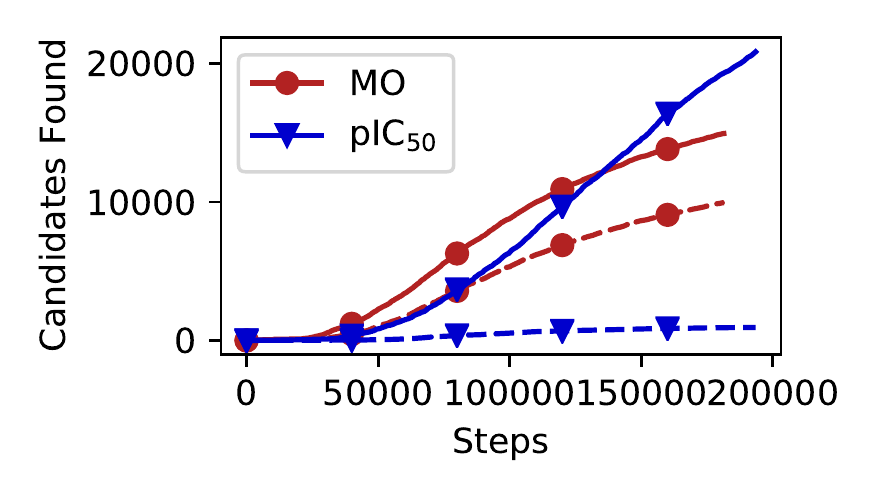}
  \caption{Performance of DQN agent using the multi-objective (MO) and pIC\textsubscript{50} reward functions. The solid lines indicate the number of unique molecules with p\icfifty > 8 found after a certain number of steps. The dashed lines indicate unique molecules with p\icfifty > 8 and QED > 0.5}
  \label{fig:rl_traject}
\end{figure*}

\begin{figure*}[htbp!] \centering
  \includegraphics[width=0.55\textwidth]{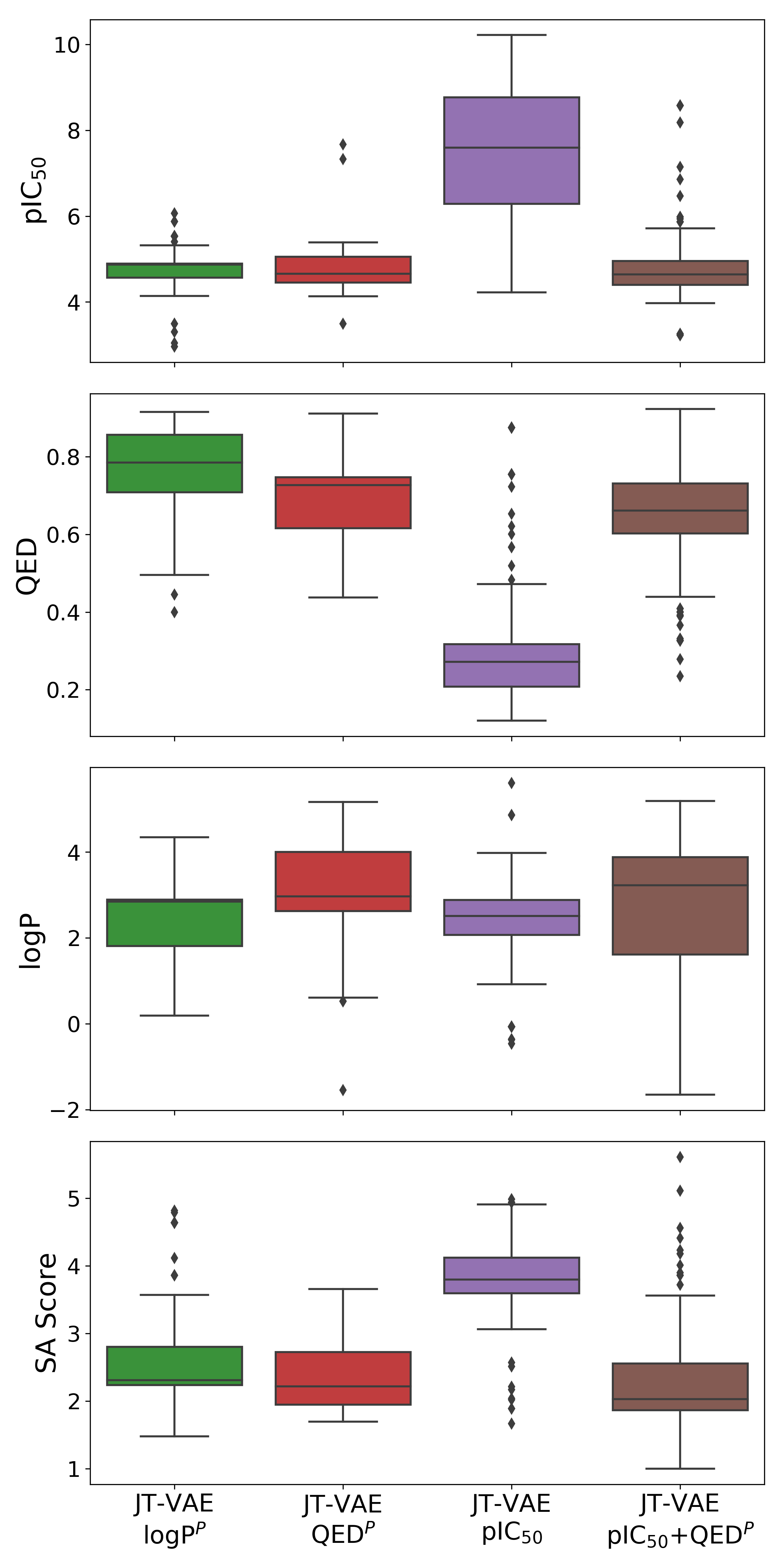}
  \caption{Comparison of the properties of molecules optimized with different scoring functions by JT-VAE method. The pIC\textsubscript{50} value was estimated using our MPNN.} \label{fig:JTVAEcomparison}
\end{figure*}

%\begin{figure*}[htbp!] \centering
%  \includegraphics[width=0.75\textwidth]{./figs/JTAVEcomparison.png}
%  \caption{Distribution of the maximum Tanimoto similarity of generated molecules compared to molecules in the database for molecules generated by each scoring function.} \label{fig:similarity}
%\end{figure*}

% \subsection{DTBA model}
% The Drug Target Binding Affinity (DTBA) model uses a drug encoding of ECFP for the drug molecule and a target encoding of Convolution Neural Network (CNN).  The target used for the validation is the  SARS-CoV2-3CL Protease\cite{chen2020prediction} with the sequence below \\

% \seqsplit{%
% SGFKKLVSPSSAVEKCIVSVSYRGNNLNGLWLGDSIYCPRHVLGKFSGDQWGDVLNLANNHEFEVVTQNGVTLNVVSRRLKGAVLILQTAVANAETPKYKFVKANCGDSFTIACSYGGTVIGLYPVTMRSNGTIRASFLAGACGSVGFNIEKGVVNFFYMHHLELPNALHTGTDLMGEFYGGYVDEEVAQRVPPDNLVTNNIVAWLYAAIISVKESSFSQPKWLESTTVSIEDYNRWASDNGFTPFSTSTAITKLSAITGVDVCKLLRTIMVKSAQWGSDPILGQYNFEDELTPESVFNQVGGVRLQ}

% The default hyperparamters supplied with the DeepPurpose were found to be sufficient. Out of the top 6 molecules generated by the JT-VAE and  validated by the JAK2 classifier as active, the DBTA model classfied  two  of those top 6 molecules have probability $\gt$ 0.5 to have interaction with SARS-CoV2-3CL Protease.

%\subsubsection{Janus kinase 2 (JAK2)}
\hfill \break 
%In order to independently validate the generated molecule we chose a well known biological target namely Janus kinase 2 (JAK2). For the binary classification a pIC$_{50}$  $\ge$ 8 is chosen as active~\cite{renz2020}. The extended connectivity fingerprint~\cite{rogers2010extended} with a length of  2048 bits and radius of 2 and a random forest classifier~\cite{breiman2001random} as implemented in scikit-learn are employed. The classifier  confirmed that the top 6 molecules generated by JT-VAE indeed as active. Further the top-2 candidate molecules are also validated in a Drug Target Binding Affinity classifier as implemented in the DeepPurpose toolkit~\cite{deeppurpose}. 
\end{document}